%% file: neurips_2025.tex
\documentclass{article}

 \usepackage[main, final, nonatbib]{neurips_2025}
\input{mymacros}

\usepackage[utf8]{inputenc} 
\usepackage[T1]{fontenc}    
\usepackage{hyperref}       
\usepackage{url}            
\usepackage{booktabs}       
\usepackage{amsfonts}       
\usepackage{nicefrac}       
\usepackage{microtype}      
\usepackage{xcolor}         
\usepackage{siunitx}
\usepackage[table]{xcolor}
\usepackage{colortbl}

\usepackage[natbib,style=numeric-comp,sorting=noneyearname,sortcites,maxcitenames=2,maxbibnames=99]{biblatex}
\renewbibmacro{in:}{}
\DeclareSortingTemplate{noneyearname}{
    \sort{\citeorder}
    \sort{\field{year}}
    \sort{\field{author}}
}

\usepackage[capitalize]{cleveref}
\usepackage{caption}

\newif\ifcolormode
\colormodetrue 

\newcommand{\setcolormodeoff}{\colormodefalse}

\setcolormodeoff 

\title{Diffusion-Driven Two-Stage Active Learning for Low-Budget Semantic Segmentation}

\author{
  Jeongin~Kim$^1$\quad
  Wonho~Bae$^2$\quad
  YouLee~Han$^1$\quad
  Giyeong~Oh$^3$\\
  \textbf{Youngjae~Yu}$^4$\quad
  \textbf{Danica~J.~Sutherland}$^{2,5}$\quad
  \textbf{Junhyug~Noh}$^1$\thanks{Corresponding author}\\[0.25em]
  $^1$Ewha Womans University \quad
  $^2$University of British Columbia\\
  $^3$Yonsei University \quad
  $^4$Seoul National University \quad
  $^5$Amii\\[0.15em]
  \texttt{\{jn.kim, oneglass, junhyug\}@ewha.ac.kr} \quad
  \texttt{\{whbae, dsuth\}@cs.ubc.ca}\\
  \texttt{hard2251@yonsei.ac.kr} \quad
  \texttt{youngjaeyu@snu.ac.kr}
}

\begin{document}

\maketitle
\setcounter{footnote}{0}

\begin{abstract}
Semantic segmentation demands dense pixel-level annotations, which can be prohibitively expensive -- especially under extremely constrained labeling budgets. In this paper, we address the problem of low-budget active learning for semantic segmentation by proposing a novel two-stage selection pipeline. Our approach leverages a pre-trained diffusion model to extract rich multi-scale features that capture both global structure and fine details. In the first stage, we perform a hierarchical, representation-based candidate selection by first choosing a small subset of representative pixels per image using MaxHerding, and then refining these into a diverse global pool. In the second stage, we compute an entropy‐augmented disagreement score (eDALD) over noisy multi‐scale diffusion features to capture both epistemic uncertainty and prediction confidence, selecting the most informative pixels for annotation. This decoupling of diversity and uncertainty lets us achieve high segmentation accuracy with only a tiny fraction of labeled pixels. Extensive experiments on four benchmarks (CamVid, ADE-Bed, Cityscapes, and Pascal-Context) demonstrate that our method significantly outperforms existing baselines under extreme pixel‐budget regimes. Our code is available at \url{https://github.com/jn-kim/two-stage-edald}.
\end{abstract}

\section{Introduction}
\label{sec:intro}

Semantic segmentation is a core task in computer vision: assign a class label to each pixel in an image.  
Applications abound in areas such as autonomous driving, robotics, and medical image analysis.  
Despite the impressive performance of convolutional~\citep{vgg2014simonyan2,resnet2016he} and Transformer-based models~\citep{vit2020dosovitskiy,deit2021touvron,liu2021swin}, an overarching challenge remains: \emph{how do we obtain the pixel-level annotations required to train these models effectively?}  
Large-scale datasets like Cityscapes~\citep{cordts2016cityscapes} and Pascal-Context~\citep{mottaghi2014context} require meticulous per-pixel labeling, rendering data collection \emph{extremely} costly.

Active learning (AL) aims to mitigate labeling costs by strategically selecting a subset of unlabeled data points, for example, pixels, for annotation \cite{kendall2017uncertainties, yoo2019learning, sener2017active}.  
However, traditional uncertainty-based AL methods for segmentation often select redundant pixels, since the $k$ most uncertain pixels are often near one another~\citep{ma2024breaking,mittal2023best}.
Representation-based methods do not select redundant pixels, but they tend to miss informative pixels near the boundary of objects.
Furthermore, as representation-based methods usually require computation of the pairwise similarity of candidates, it is prohibitive to consider all the available pixels at once. 

In real-world scenarios, the total number of pixels is massive; for example, $N$ images each of resolution $H \times W$ yield a total of $N\!\cdot\!H\!\cdot\!W$ pixels.
However, because of the high annotation cost, the labeling budget can be orders of magnitude smaller.  
We therefore formalize a \emph{low-budget active learning} setting for semantic segmentation: at each AL round, one may annotate only $b$ pixels from the entire unlabeled pool, with $b \ll N\!\cdot\!H\!\cdot\!W$.
In our standard (extreme) regime, we allocate \emph{one pixel per image} for $10$ rounds -- \ie $b=0.1N$ pixels per round and a total of $\approx N$ pixel labels across the whole process.

A promising direction to address this data-scarce labeling scenario is to harness the representation power of \emph{diffusion models} \cite{ho2020denoising, rombach2022high}.
Diffusion models are able to iteratively denoise over multiple time steps, generating highly detailed images~\cite{ho2020denoising}. The initial steps of this reverse diffusion process capture the overall structure of objects, while the later steps focus on generating fine-grained details.
\citet{baranchuk2022ledm} demonstrate that multi-timestamp features from diffusion models are beneficial for semi-supervised segmentation tasks.
We will show that they can be used for uncertainty estimation, due to their ensemble-like nature.

We propose a two-stage strategy that efficiently pinpoints both diverse and epistemically uncertain pixels. 
In \textbf{Stage 1}, we employ a representation-based AL method, in particular MaxHerding~\citep{bae2024maxherding}, to extract a set of candidate pixels which is manageable in size yet representative; we first narrow down candidates in each image, then refine across all images to ensure global diversity. 
In \textbf{Stage 2}, we exploit the diffusion backbone’s stochastic multi‐scale features to compute an \emph{entropy‐augmented disagreement} score (eDALD): we measure mutual information between noisy feature samples and labels, then add a single‐sample entropy term to capture prediction confidence. 
This combined criterion prioritizes pixels that both lie in under‐explored regions of feature space and carry high model uncertainty, maximizing the benefit of each annotation under an extremely limited pixel budget.

Our contributions are summarized as follows:
\begin{itemize}[leftmargin=*]
    \item We formalize and address the challenging \emph{low-budget active learning for semantic segmentation}, where a mere fraction of pixels can be annotated per round.
    \item We introduce a scalable two-stage pipeline (coverage $\to$ uncertainty) for pixel-level AL: a \emph{local-then-global} MaxHerding stage yields a representative candidate pool, followed by uncertainty-based refinement.
    \item We develop a diffusion-native uncertainty criterion (eDALD) that combines disagreement from stochastic \emph{multi-timestep} features with a single-pass entropy term; \emph{used after coverage}, it complements MaxHerding and yields substantial gains over one-stage uncertainty-only or coverage-only variants under tiny budgets.
    \item We present \emph{comprehensive experiments} on benchmark datasets -- CamVid, ADE-Bed, Cityscapes, and Pascal-Context -- demonstrating consistent gains over multiple baselines when budgets are severely constrained.
\end{itemize}


\section{Related Work}
\label{sec:related}

\subsection{Active Learning for Classification}
Active learning (AL) is a learning framework to improve the data efficiency of training machine learning models by strategically selecting the most informative data points for annotation. 
AL methods can be broadly categorized into two main approaches: uncertainty-based methods and representation-based methods.

\noindent\textbf{Uncertainty-based Methods.} 
Uncertainty-based methods~\citep{entropy2014wang,margin2001scheffer,sequential1994lewis,heterogeneous1994lewis} generally select data points near the decision boundary.
These simple methods are generally effective; however, they can be prone to selecting pixels with erratic or extreme predicted class probabilities. These pixels, such as isolated noise artifacts or very rare patterns, often exhibit high uncertainty yet lie far from the true decision boundary, contributing little to model improvement.
To avoid this problem, some work has attempted to measure the change of a model for a given candidate data point and its corresponding pseudo-label~\citep{al2009settles,egl2007settles,badge2019ash} or the change of model outputs~\citep{emoc2014frey,emoc2016kading,emoc_reg2018kading,lookahead2022mohamadi};
in practice, these methods typically do not perform much better than the simple ones when using predictors defined by deep networks.

From a Bayesian perspective, on the other hand, the goal of AL is essentially to select data points that maximize mutual information (or equivalently, information gain) between model parameters $\theta$ and the true label $Y$ to be observed, given an input $x$ and a model $f$.
\begin{align}
    \tilde{\x}^* &\in \argmax_{{\x} \in \setU} \funcI(\theta; Y \:|\: \x)\ \text{ where }\ \funcI(\theta; Y \:|\: \x) =  \funcH(Y \mid \x) - \E_{\theta} \Bigl[ \funcH(Y \mid \theta, \x) \Bigr]
    \label{eq:bald_select}
\end{align}
This objective is known as Bayesian Active Learning by Disagreement (BALD), since it selects a data point where model parameters under the posterior distribution disagree the most for the output label~\citep{houlsby2011bald}.  
PowerBALD~\citep{kirsch2021powerbald} avoids redundant top-\(k\) selection by sampling without replacement from a ``powered'' BALD distribution \(p(i)\propto s_i^\beta\).  
Balanced Entropy (BalEnt) and its acquisition variant BalEntAcq~\citep{woo2021balent} fit each softmax marginal to a Beta distribution, producing a bounded score that balances epistemic and aleatoric uncertainty.

\noindent\textbf{Representation-based Methods.}
These methods instead select samples that are representative of the entire unlabeled data pool, often by analyzing learned data embeddings or features~\citep{mittal2023best,xie2020deal}. Core-set methods, for example, aim to find a small subset of the unlabeled data that best covers the diversity of the entire dataset in a feature space~\citep{xie2020deal,lyu2024semi,sener2017active}. Another notable approach is Variational Adversarial Active Learning (VAAL), which employs a variational autoencoder and a discriminator to identify samples that are most distinct from the already labeled data in a learned latent space~\citep{lyu2024semi,sinha2019variational}.

\citet{typiclust2022hacohen} recently demonstrated that representation-based methods perform significantly better than uncertainty-based methods in low-budget regimes. 
A sequence of works has proposed improving notions of coverage to select better representing samples: probability coverage~\citep{probcover2022yehuda}, generalized coverage~\citep{bae2024maxherding} and uncertainty coverage~\citep{uherding2024base}.

\subsection{Active Learning for Semantic Segmentation}

Active learning for semantic segmentation extends classification -- style AL to the pixel‐wise labeling domain, where annotation costs are dramatically higher.
Unlike classification, every pixel’s label matters, so AL methods must balance selection granularity, spatial coherence, and computational efficiency.
Broadly, existing approaches fall into three categories -- image‐level, region‐level, and pixel‐level -- each trading off annotation cost against precision and implementation complexity.

\noindent\textbf{Image-level Methods.}
The most straightforward adaptation of classification‐style active learning to segmentation is to select entire images for annotation.  
Many image-level selection strategies for semantic segmentation directly adapt techniques from classification.
For instance, uncertainty-based methods often calculate a per-pixel uncertainty score and then aggregate these scores across the entire image, such as by averaging, to decide which images to query~\citep{mittal2023best}.
\citet{xie2020deal} also employ image-level selection, though they enhance it by considering semantic difficulty.
Similarly, representation-based approaches like Core-set or VAAL can be applied by selecting images whose overall feature representations contribute most to dataset diversity or are most distinct from already labeled data~\citep{xie2020deal,sinha2019variational}.
While simple to implement, image‐level querying quickly exhausts annotation budgets because every pixel -- even in well‐understood regions -- must be annotated.

\noindent\textbf{Region-level Methods.}  
To reduce per‐query cost, region‐level approaches annotate clusters of pixels -- such as superpixels~\citep{cai2021revisiting, kim2023adaptive}, rectangular patches~\citep{casanova2020reinforced}, or bounding boxes~\citep{kasarla2019region}.
These methods save clicks by labeling chunks at once and often combine uncertainty with coverage or learned difficulty.
However, they assume a minimally competent model to estimate region scores: under extreme low budgets (``cold‐start''), unreliable uncertainty or difficulty estimates can misrank regions and degrade performance.
Moreover, multi‐class regions can introduce label ambiguity when boundaries cross cluster edges.

\noindent\textbf{Pixel-level Methods.}  
Pixel‐level querying -- selecting individual pixels -- incurs the lowest per‐unit cost and avoids region‐boundary ambiguity but has received less attention, as each query yields limited information.  
PixelPick~\citep{pixelpick2021shin} applies margin sampling to individual pixels, and \citet{didari2024bayesian} use BalEntAcq~\citep{woo2021balent} as a Bayesian pixel‐uncertainty measure.
Purely uncertainty‐driven pixel selection, however, demands many queries and often fails in low‐budget regimes.

In contrast to these prior works, we introduce the first practical framework for \emph{low‐budget} active learning in semantic segmentation.
Our two-stage pipeline first uses representation-based sampling to build a compact, diverse candidate pool, then applies uncertainty-driven selection to pick the final pixels -- achieving high segmentation accuracy with only a handful of annotated points in low-budget scenarios.

\subsection{Diffusion Models in Vision}

\noindent\textbf{Generative Diffusion.}
Denoising Diffusion Probabilistic Models (DDPM)~\citep{ho2020denoising} and score SDE~\citep{song2019generative,song2020improved,song2020score} introduce a novel paradigm of generative models: progressively destroy an image by adding noise, then learn to gradually reverse this process. 
Follow-up works such as Latent Diffusion Models (LDM) \cite{rombach2022high} improve scalability for high-resolution images.

\noindent\textbf{Diffusion for Downstream Tasks.}
Some recent research explore leveraging the representations from diffusion models for tasks beyond generation \cite{ji2023ddp}. 
In particular, medical image segmentation \cite{wu2024medsegdiff} and few-shot segmentation \cite{baranchuk2022ledm} benefit from multi-scale features spanning from broad global context (earlier denoising steps) to detailed object boundaries (later steps). 
Nevertheless, the integration of diffusion representations into \emph{active learning} remains unexplored.

\begin{figure*}[t!]
\centering
\includegraphics[width=1.00\textwidth]{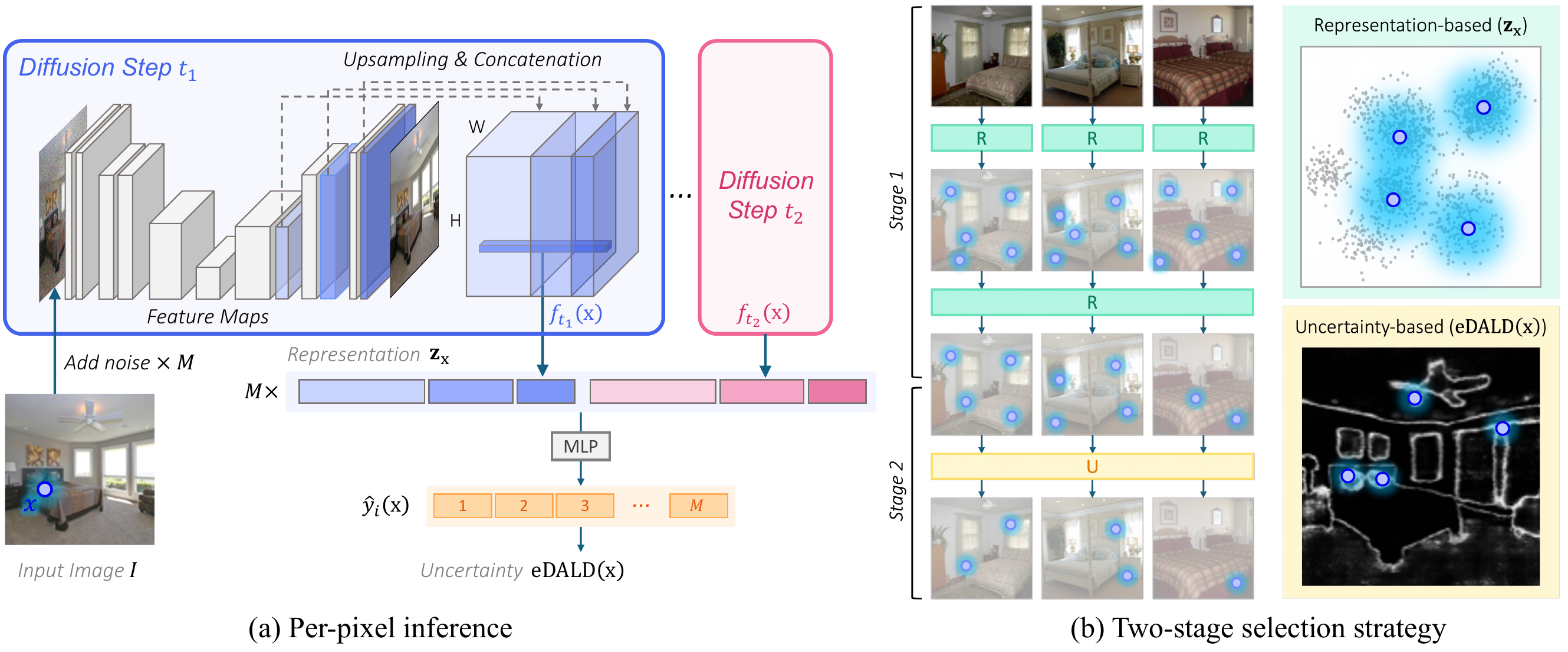}
\caption{
Overview of the proposed pipeline (illustrative hyperparameters).  
(a) For each pixel \(x\), the pre‐trained diffusion model yields a feature representation \(\vecz_x\), an uncertainty estimate \(\mathrm{eDALD}(x)\), and the class prediction \(\hat y(x)\) (here shown with \(T=2\) diffusion steps and \(L=3\) layers purely for illustration).  
(b) These outputs drive our two‐stage selection: Stage 1 picks \(K=5\) representative pixels per image to form a global pool of size \(M=10\), then Stage 2 applies the uncertainty criterion to select the final \(b=5\) pixels (all parameter values are example settings, not the exact experimental hyperparameters).
}
\label{fig:overview}
\end{figure*}

\section{Proposed Method}
\label{sec:proposed}

In this work, we present a novel two-stage active learning pipeline tailored for semantic segmentation under a strictly low-budget setting. 
Our approach integrates the strengths of pre-trained diffusion models for feature extraction with a two-pronged sample selection strategy that first ensures representation diversity and then refines the selection using uncertainty estimation. 
The overall pipeline is illustrated in Figure~\ref{fig:overview} and detailed in Algorithm~\ref{alg:two_stage_al}.
             

\subsection{Preliminaries}
\label{sec:background}

\paragraph{Problem Setup.}
Let $\,\setI = \{I^{(n)}\}_{n=1}^N$ denote a set of $N$ images, where each image $I_i$ contains $H \times W$ pixels. 
At each AL round $r \in [R]$ with a labeled pixel set $\setL_r$ and an unlabeled set $\setU_r$, we select a subset $\setS_r \subseteq \setU_r$ such that $|\setS_r| = b$ for annotation, where $b$ is a pre-defined labeling budget per round.
More specifically, active learning involves the following steps in each $r$-th round:
\begin{enumerate}
    \item Measure how informative each pixel in (or a subset of) $\mathcal{U}_r$ is
    \item Select a subset $\setS_r \subseteq \setU_r$ that consists of $b$ ``most informative'' pixels
    \item Update both $\setL_{r+1} \leftarrow \setL_r \cup \setS_r$ and $\setU_{r+1} \leftarrow \setU_r \setminus \setS_r$
    \item Retrain the segmentation model on $\setL_{r+1}$
\end{enumerate}

In this work, we focus on extremely low-budget setting \ie $b \ll (N \times H \times W)$, the total number of pixels.
The challenge is then to select highly informative pixels for robust generalization in semantic segmentation tasks. 


\begin{algorithm*}[t!]
\caption{Two-Stage Low-Budget Active Learning at Round $r$} 
\label{alg:two_stage_al}
\begin{algorithmic}[1]
\Require 
  Image set $\mathcal{I}$, 
  labeled pixel set $\setL_r$,
  unlabeled pixel set $\setU_r$,
  diffusion feature extractor $f$, 
  current segmentation head $\seghead$, 
  the number of candidates per image $K$, 
  global candidate pool size $M$, 
  per-round budget $b$.
\Ensure Update labeled set $\setL_r$ and unlabeled set $\setU_r$ with $b$ new annotated pixels.
\medskip

\State $\mathcal{M}_0 \gets \emptyset$  \textcolor{blue}{\texttt{\: // Initialize the global candidate pool}}

\State \textcolor{blue}{\texttt{// Stage 1: Representation-based Candidate Selection}}
\For{each image $I \in \mathcal{I}$}
  \State Extract multi-scale features $\mathcal{Z} = \{\vecz_\x\}_{\x \in I, \x \in \mathcal{U}}$ using $f$
  \State Select a set of $K$ representative pixels $\setR$ by applying MaxHerding to $\mathcal{Z}$
  \State $\setM_0 \gets \setM_0 \cup \setR$
\EndFor
\State Obtain the global candidate pool $\setM$ of size $M$ by applying MaxHerding to $\setM_0$

\State \textcolor{blue}{\texttt{// Stage 2: Uncertainty-driven Selection}}


\For{each $\x \in \mathcal{M}$}
  \State Compute $\mathrm{eDALD}(\x)$ as in Eq.~\eqref{eq:edald_def} using the segmentation head $s_\theta$ and $f$
\EndFor

\State $\mathcal{B}_r \gets$ Top-$b$ pixels from $\mathcal{M}$ sorted by descending $\mathrm{eDALD}(\x)$
\State $\setL_{r+1} \gets \setL_r \cup \setB_r,\, \setU_{r+1} \gets \setU_r \setminus \setB_r $
\State \Return $\setL_{r+1}$, $\setU_{r+1}$
\end{algorithmic}
\end{algorithm*}


\paragraph{Diffusion-based Semantic Segmentation.}
Our segmentation framework builds on the LEDM architecture~\citep{baranchuk2022ledm} and exploits a pre-trained diffusion model denoted as $f(\cdot)$ -- \eg DDPM~\citep{ho2020denoising} to extract robust multi-scale features.
Note that it is crucial to utilize good representations, particularly in low-budget regimes, as a segmentation model trained on a small labeled set itself would not be sufficient to learn good representations.

For an image $I$, the diffusion model produces feature maps at multiple denoising timesteps $t \in \{t_1, t_2, \ldots, t_T\}$ and layers $l \in \{l_1, l_2, \ldots, l_L\}$.
For a given pixel $\x$ in an image $I$ (at coordinates $(w,h)$, where $1\leq w \leq W$ and $1\leq h \leq H$), let $f_{t,l}(\x) \in \mathbb{R}^{D_{t,l}}$ denote a feature vector from a layer $l$ at denoising timestep $t$.
We obtain a \emph{multi-scale} representation by concatenating these features:
\begin{equation}
\begin{split}
\vecz_\x \vcentcolon= \Big[\, f_{t_1,l_1}(\x); \ldots; f_{t_1,l_L}(\x); 
            \ldots; f_{t_T,l_1}(\x); \ldots; f_{t_T,l_L}(\x) \,\Big] \in \mathbb{R}^{D},
\end{split}
\label{eq:diff_feature}
\end{equation}
where $D=\sum_{i=1}^{T}\sum_{j=1}^{L}D_{t_i,l_j}$. This comprehensive feature vector $\vecz_\x$ encapsulates rich semantic cues learned during the diffusion process as demonstrated in \citep{baranchuk2022ledm}. 
We omit the subscript $\x$ when it is clear from context. 

The predicted probability of segmentation for $C$ object classes is produced by a lightweight segmentation head $\seghead:\mathbb{R}^D \rightarrow \mathbb{R}^C$.
It consists of a 2-layer MLP with a ReLU activation and batch normalization, followed by a softmax layer. 
Note that we deliberately add a parameter notation $\theta$ only for the segmentation head $\seghead$ not for the diffusion model $f$, to describe that we only update the segmentation head not the diffusion model. 


\subsection{Representation-based Candidate Selection}
\label{subsec:stage1}

Representation-based AL methods rely on the pairwise similarity measure.
Hence, it is often infeasible to select pixels from all possible $(N\times H \times W)$ pixels, as it requires $O(N^2W^2H^2)$ computation for a pairwise comparison.
To tackle this problem, we adopt a \emph{two-step} strategy that leverages the MaxHerding algorithm~\cite{bae2024maxherding}, a representation-based AL method, in each step to ensure diversity while reducing computational complexity; please refer to Line $1$--$8$ in \cref{alg:two_stage_al}.

We first identify $K$ representative pixels \emph{within each image} $I$.
More specifically, for each pixel $\x \in I$, we obtain a multi-scale feature vector $\vecz$ extracted from the diffusion model $f$ using Eq.~\eqref{eq:diff_feature}.
We then greedily select an optimal pixel $\tilde{\x}^*$ at a time using generalized coverage $\hat{\mathrm{C}}_k$~\citep{bae2024maxherding} as follows:
\begin{align}
    \tilde{\x}^* \in \argmax_{\tilde{\x} \in I, \tilde{\x} \in \setU} \hat{\mathrm{C}}_k(\setL \cup \{\tilde{\x} \})\ \text{ where }\ \hat{\mathrm{C}}_k(\setL \cup \{\tilde{\x}\}) \coloneqq \frac{1}{\lvert\setU \rvert} \sum_{\x \in \setU} \left[ \max_{\x'\in\setL \cup \{\tilde{x}\}} k(\x, \x') \right].
    \label{eq:gcover_select}
\end{align}
Here, $k$ is a function that measures pairwise similarity between $\x$ and $\x'$.
We simply use a RBF kernel for $k$ as
$k(\x, \x') = \exp\left( -\frac{\lVert \vecz_\x- \vecz_{\x'} \rVert^2_2}{\sigma^2} \right)$.
We repeat this until we obtain $K$ pixels yielding a compact yet diverse subset $\setR$ for each image $I$.

By restricting the search space for selecting pixels to an individual image, we reduce the size of candidates to $(H \times W)$ compared to $(N \times H \times W)$. 
We obtain the initial global pool $\setM_0$ by merging the selected subset $\setR$ for each image $I$.
This pool has size $(N \times K)$, which is much smaller than $(N\times H \times W)$.
To further refine these candidates, we apply MaxHerding again \emph{across the entire merged set} to select $M$ representative pixels, which form a candidate pool for the final $b$ pixels.
The final global pool $\setM$ of size $M$ thus achieves good coverage over all images collectively.
Typically, $M \ll N\cdot H\cdot W$.

Through this stage, we ensure that our final candidate set $\mathcal{M}$ is both \emph{diverse} (capturing the data manifold effectively) and \emph{manageable in size}, laying the foundation for more targeted selection via uncertainty in the next step.


\subsection{Uncertainty-driven Selection}
\label{subsec:stage2}

\paragraph{Diffusion-based Active Learning by Disagreement (DALD).} From the diverse candidate pool $\mathcal{M}$, we further refine our selection by identifying the most informative $b$ pixels using an uncertainty-based measure; refer to Line $9$--$13$ in \cref{alg:two_stage_al}. 
Our method uniquely exploits the structure of the diffusion model: each pixel $\x$ is associated with multiple feature vectors.

Inspired by the BALD objective in \cref{eq:bald_select}, we propose \textit{Diffusion-based Active Learning by Disagreement} (DALD) that selects a new data point as follows:
\begin{align}
    \x^* \in \argmax_{\x \in \setU}\ \funcI(\hat{Y}; Z \mid \x, \seghead, f)
    \label{eq:our_bald_select}
\end{align}
where $\hat{Y}$ denotes a random variable for the predicted label of a pixel $\x$ and $Z$ denotes a random variable for the concatenated multi-scale features from $\x$ computed using \cref{eq:diff_feature}.
Note that the stochasticity of $Z$ comes from noise added to $\x$ at $t_1,\dots, t_T$ timesteps.
We define this conditional distribution of $Z$ given $\x$ as $q(\cdot \mid \x; f)$. 
Furthermore, suppose $X$ denotes a random variable for a pixel $\x$, then the computational probabilistic graphical model is defined as $X \rightarrow Z \rightarrow \hat{Y}$, assuming that $X$ does not provide additional information about $\hat{Y}$ knowing $Z$.


The motivation of DALD is to select a data point where information gain for a noised multi-scale feature $Z$ is maximized given a label $y$.
Intuitively, if the information gain of a multi-scale feature is close to 0, knowing the label $y$ does not help removing the ambiguity of the feature,\footnote{It can happen either when aleatoric uncertainty is very high or epistemic uncertainty is very low.} while if the information gain of a multi-scale is high, knowing the label $y$ significantly removes the uncertainty of the noised feature.

The mutual information in \cref{eq:our_bald_select} can be decomposed into the following two terms:
\begin{align}
    \funcI(\hat{Y}; Z \mid \x, \seghead, f) = \underbrace{ \vphantom{\E_{\x'}} \funcH(\hat{Y} \mid \x, s_\theta, f)}_{\text{Unconditional entropy}}  - \E_{\vecz \sim q(\cdot \mid \x) }   \biggl[\  \underbrace{ \vphantom{\E_{\x'}} \funcH(\hat{Y} \mid Z=\vecz, \x, s_\theta) }_{\text{Conditional entropy}}\ \biggr]. 
    \label{eq:our_bald}
\end{align}
We compute the uncertainty of the predicted label $\hat{Y}$ instead of a multi-scale feature $Z$ to make computation easy.

The computation of conditional entropy is relatively more straightforward.
As $\hat{Y}$ is conditionally independent to a clean pixel $\x$ given a multi-scale feature $\vecz$, the expected conditional entropy can be approximated using $M$ samples of $\vecz$, as follows:
\begin{align}
   \E_{\vecz \sim q(\cdot \mid \x) } \bigl[ \funcH(\hat{Y} \mid Z=\vecz, \x) \bigr] \approx -\frac{1}{M}\sum_{m=1}^M \sum_{y \in \setY} \hat{p}_\theta (y|\vecz^{(m)})\cdot \log{ \hat{p}_\theta (y|\vecz^{(m)})}
\end{align}
where $\hat{p}_\theta (y|\vecz) = \seghead (\vecz)_y$.
Therefore, conditional entropy is approximated as the mean of entropy.
In contrast, we compute the unconditional entropy as the entropy of the mean predictions as follows:
\begin{align}
   \funcH(\hat{Y} \mid \x, s_\theta, f) \approx -\sum_{y \in \setY} \bar{p}_\theta (y|\x) \cdot \log{ \bar{p}_\theta (y|\x)}, \text{ where }\ \bar{p}_\theta (y|\x) = \frac{1}{M}\sum_{m=1}^M \hat{p}_\theta(y | \vecz^{(m)}).
\end{align}

\paragraph{Entropy‐Augmented DALD (eDALD).}
Although disagreement-based selection has the advantage of being less sensitive to aleatoric (or irreducible) uncertainty, it also has a clear limitation: it does not account for the confidence of model predictions.
For example, two different segmentation models may output identical mutual information for a given pixel, although one may produce a low entropy (high confidence) output, while the other yields a high entropy (low confidence) output.
This discrepancy illustrates that disagreement alone may fail to capture the full picture of predictive uncertainty.

As a simple and computationally inexpensive remedy to this limitation, we introduce an additional entropy term based on a separate sample.
Specifically, our selection objective becomes the following:
\begin{align}
\x^* &= \argmax_{\x\in\mathcal U}\;\mathrm{eDALD}(\x), \text{ where }\
\mathrm{eDALD}(\x) = \funcI \bigl(\hat{Y} ;Z \mid \x, s_\theta, f\bigr)
   + \funcH \bigl(\hat{Y} \mid \vecz^{(0)}, \x\bigr).
\label{eq:edald_def}
\end{align}
where $\vecz^{(0)}$ denotes an independently drawn sample, separate from sample $m=1,2,\dots,M$ used to estimate the mutual information.
This extra entropy term highlights pixels where the model is less confident, further enhancing the sensitivity of acquisition to both disagreement and absolute uncertainty.




\subsection{Overall Selection and Training Procedure}
\label{subsec:training}

Our complete active learning pipeline is summarized in Algorithm~\ref{alg:two_stage_al}. 
At each of the \(R\) rounds, we (1) extract multi‐scale diffusion features $\vecz_\x$ for all unlabeled pixels $\x \in \setU$, (2) select \(b\) pixels through our two‐stage sampler (MaxHerding $\to$ eDALD), (3) annotate their labels and update the labeled/unlabeled sets, and (4) train on the expanded labeled set.

During training, the segmentation head $\seghead$ is optimized via a cross-entropy loss computed over the up-to-date labeled set $\mathcal{L}$:
\[
    \theta^* \in \argmin_{\theta} - \frac{1}{\lvert \mathcal{L} \rvert}\sum_{(\x, y) \in \mathcal{L}}  \log \hat{p}_\theta(y\,\lvert\, \x, f).
\]

This cycle is repeated over $R$ rounds, allowing the model to progressively learn from a carefully curated and highly informative set of pixels while keeping the annotation costs minimal.

\section{Experiments}
\label{sec:experiments}

\subsection{Datasets and Setup}
We evaluate our low‐budget active learning pipeline on four standard semantic segmentation benchmarks. 
All images are processed at $256\times256$ resolution for compatibility with the diffusion backbone:
\begin{itemize}[leftmargin=*]
  \item \textbf{CamVid}~\cite{brostow2009semantic}: An urban driving dataset with $367$ train and $233$ test images, each labeled into 11 classes. All images are center‐cropped to $256\times256$.
  \item \textbf{ADE-Bed:} A subset of ADE20K~\cite{zhou2017scene} consisting of bedroom images with $964$ train and $650$ test images annotated with the $30$ most common object classes. We resize the shorter side to $256$ pixels, preserve aspect ratio, then center-crop to $256\times256$.
  \item \textbf{Cityscapes}~\cite{cordts2016cityscapes}: A street‐scene dataset comprising $2{,}975$ train and $500$ validation images over $19$ classes. We resize and center‐crop to $256\times256$.
  \item \textbf{Pascal‐Context}~\citep{mottaghi2014context}: A scene parsing dataset providing dense semantic labels for more than 400 categories. Following convention~\citep{mottaghi2014context,bansal2017pixelnet,gu2020cagnet}, we use $33$ most frequent categories. It contains $4{,}998$ train and $5{,}105$ validation images. Images are resized to $256\times256$ using bilinear interpolation.
\end{itemize}

\paragraph{Budget Setting.}
We fix a total annotation budget $B$ to be the average of 1 labeled pixels per image, \ie $B = N$.
For Pascal‐Context with $N\!=\!5{,}000$, this implies $B=5{,}000$ pixels -- a practical scale for large datasets.
This budget is evenly split across $R=10$ AL rounds, yielding $b = B/R = 0.1N$ pixels per round.
As intended, we only annotate $0.0015\%$ of the total pixels after 10 AL rounds.

\subsection{Implementation Details}

\paragraph{Diffusion Model.}
We adopt an ImageNet pre-trained diffusion model~\cite{ho2020denoising} for feature extraction from a publicly available \href{https://github.com/openai/guided-diffusion.git}{guided diffusion repo}.
 To capture multi-scale cues, we sample $T=3$ timesteps ($t_1=50, t_2=150, t_3=250$). At each timestep, we extract the features from $L=4$ layers ($l_1=5, l_2=8, l_3=12, l_4=17$), resulting in a rich, concatenated representation for each pixel.
For uncertainty estimation (DALD/eDALD), we draw $M=5$ noisy feature samples per pixel.

\paragraph{Candidate Selection.}
Following Section~\ref{subsec:stage1}, we apply MaxHerding~\citep{bae2024maxherding} to the pixels of each image to obtain $K=50$ representative samples, which are then merged across all images into an initial global pool $\setM_0$.
We further apply MaxHerding to $\setM_0$ to obtain the final candidate set $\setM$, which contains half of the samples in $\setM_0$.

\paragraph{Training Procedure.}
We use Adam with an initial learning rate of $1\times10^{-3}$ for training the pixel classifier, with a batch size of $5$.
We apply early stopping if the segmentation loss does not improve for $50$ consecutive iterations and the target pixel accuracy exceeds $95\%$.
This training procedure is repeated for $10$ active learning rounds, each time adding $b$ newly labeled pixels to the training set.


\subsection{Quantitative Results}
\label{sec:quant_results}

\paragraph{Representation‐First \vs\ Uncertainty‐Only.} 
All results in Table~\ref{tab:exp_stage} use our diffusion‐backbone on CamVid dataset. We compare ``UC Only’’ (single‐stage: uncertainty) against ``Herding $\to$ UC'' (two‐stage: representation $\to$ uncertainty), reporting absolute and relative gains.
Overall, enforcing diversity first consistently boosts simple criteria -- entropy gains $+5.51$ mIoU ($+21.81\%$), and margin sampling gains $+1.5$ mIoU ($+4.8\%$) -- confirming that representation filtering improves even basic uncertainty sampling.  

However, pure DALD (random‐noise) and BALD (MC‐Dropout) actually worsen after MaxHerding ($-2.76$ and $-1.8$ mIoU, respectively), suggesting that disagreement‐only measures can overemphasize noisy or redundant regions once diversity is already enforced. In contrast, their entropy‐augmented versions show dramatic improvements: eBALD gains $+6.16$ mIoU ($+23.73\%$), and eDALD gains $+10.98$ mIoU ($+43.68\%$). 
This striking boost reflects the complementary roles of the two signals: disagreement identifies perturbation‐sensitive pixels that entropy alone may overlook, while entropy recovers consistently low‐confidence areas that disagreement misses. Their combination reshapes the acquisition ranking rather than merely scaling one criterion, leading to more balanced and informative selections.
Eventually, ``Herding $\to$ eDALD'' achieves the highest mIoU of $36.12$, demonstrating that combining disagreement with confidence‐aware uncertainty, along with representation diversity, is crucial to achieve substantial gains under extreme low‐budget settings.

\begin{table}[htbp]
\centering
\newcommand{\pms}[1]{~$\pm$~#1}

\caption{
    Effect of representation-first filtering on uncertainty sampling on CamVid, measured in mIoU (\%). ``UC Only'' shows single‐stage performance; ``Herding $\to$ UC'' adds MaxHerding before uncertainty. Gains are shown in absolute points and percentages, with positive values in blue and negative in red.
    All results are based on the mean $\pm$ std from three independent runs.
}
\label{tab:exp_stage}
\small
\setlength{\tabcolsep}{10pt}
{
\begin{tabular}{c|cc|cc}
\toprule
\textbf{Uncertainty} & \textbf{UC Only} & \textbf{Herding $\to$ UC} & \textbf{Gain (pp)} & \textbf{Gain (\(\boldsymbol{\%}\))} \\
\midrule
Entropy         & 25.26\pms{0.36} & 30.77\pms{0.44} & \blue{$+$5.51} & \blue{$+$21.81} \\
Margin & 31.27\pms{1.10} & 32.77\pms{0.75} & \blue{$+$1.50} & \blue{$+$4.80}  \\
BALD            & 24.59\pms{0.97} & 22.79\pms{0.89} & \red{$-$1.80} & \red{$-$7.32}  \\
DALD            & 23.81\pms{3.60} & 21.05\pms{1.05} & \red{$-$2.76} & \red{$-$11.59} \\
PowerBALD       & 30.03\pms{0.76} & 31.57\pms{0.79} & \blue{$+$1.54} & \blue{$+$5.13} \\
PowerDALD       & 31.30\pms{1.22} & 32.00\pms{0.66} & \blue{$+$0.70} & \blue{$+$2.24}  \\
eBALD (Entropy + BALD)  & 25.96\pms{1.92} & 32.12\pms{0.40} & \blue{$+$6.16} & \blue{$+$23.73} \\
eDALD (Entropy + DALD)  & 25.14\pms{0.57} & 36.12\pms{0.24} & \blue{$+$10.98}& \blue{$+$43.68} \\
\bottomrule
\end{tabular}
}
\end{table}

\paragraph{Performance Comparison with Baselines.}
Table~\ref{tab:al_results} compares mIoU after $10$ rounds under a severe pixel‐budget across four datasets.
Pixel‐level AL research is still nascent, so we benchmark against two representative methods -- PixelPick~\cite{pixelpick2021shin} and \citet{didari2024bayesian} -- both originally designed for settings where a fixed number of pixels per image can be labeled each round.
In our extreme low‐budget regime (on average only $0.1$ pixels per image per round are labeled), we select the top-$b$ most uncertain pixels from the unified candidate pool across the dataset according to the acquisition criterion.

Given PixelPick is equivalent to Margin selection with DeepLabV3 backbone, the benefit of DDPM-backbone over DeepLabV3 can be estimated by comparing PixelPick (1st row) \vs\ Margin (5th row): DDPM-backbone improves the performance by $11.81$ mIoU on average ($22.85$ \vs\ $34.66$), and by up to $21.68$ mIoU on the ADE-Bed dataset ($8.35$ \vs\ $30.03$).
Similarly, \citet{didari2024bayesian} (BalEntAcq + DeepLabV3, 2nd row) generally performs worse than BalEntAcq with DDPM-backbone (6th row).
Putting all together, our proposed method, 2-Stage eDALD, achieves the best overall results, consistently outperforming all baselines on average by non-trivial margins (from $2.48$ to $17.36$ mIoU).
This demonstrates that the entropy-augmented diffusion-based mutual information within a diverse candidate pool is notably effective under extreme low-budget regimes.


\begin{table}[htbp] 
\centering 
\newcommand{\pms}[1]{~$\pm$~#1} 
\caption{ 
mIoU (\%) of active learning methods under a low-budget regime (10 rounds). Our two-stage method is highlighted in gray.
All results are based on the mean $\pm$ std from three independent runs.
} 
\label{tab:al_results} 
\small 
\setlength{\tabcolsep}{5pt} 
{
\begin{tabular}{cl|cccc|c}
\toprule
\textbf{Backbone} & \textbf{Method} & \textbf{CamVid} & \textbf{ADE-Bed} & \textbf{Cityscapes} & \textbf{Pascal-C} & \textbf{Avg} \\
\midrule
\multirow{2}{*}{DeepLabV3~\cite{chen2017deeplabv3}}
  & PixelPick~\cite{pixelpick2021shin} & 29.93\pms{0.12} & 8.35\pms{0.41} & 26.82\pms{0.14} & 26.28\pms{0.09} & 22.85 \\
  & \citet{didari2024bayesian} & 22.47\pms{0.10} & 8.66\pms{0.53} & 19.85\pms{0.07} & 28.15\pms{0.11} & 19.78 \\
\midrule
\multirow{7}{*}{DDPM~\cite{ho2020denoising}}
  & Random      & 25.91\pms{1.23} & 17.83\pms{0.62} & 27.13\pms{1.38} & 41.70\pms{2.08} & 28.14 \\
  & Entropy     & 25.26\pms{0.36} & 23.02\pms{1.64} & 28.62\pms{1.05} & 42.09\pms{1.99} & 29.74 \\
  & Margin      & 31.27\pms{1.10} & 30.03\pms{0.37} & 32.23\pms{1.21} & 45.11\pms{2.45} & 34.66 \\
  & BalEntAcq   & 19.37\pms{1.10} & 17.48\pms{1.36} & 24.04\pms{2.07} & 33.06\pms{4.18} & 23.49 \\
  & eDALD       & 25.14\pms{0.57} & 23.06\pms{1.29} & 29.44\pms{1.38} & 43.05\pms{0.12} & 30.17 \\
  & \cellcolor{gray!20}2-Stage eDALD & \cellcolor{gray!20}\bf{36.12}\pms{0.24} & \cellcolor{gray!20}\bf{31.12}\pms{0.20} & \cellcolor{gray!20}\bf{33.34}\pms{0.78} & \cellcolor{gray!20}\bf{47.98}\pms{0.41} & \cellcolor{gray!20}\bf{37.14} \\
\bottomrule
\end{tabular}
}
\end{table}

\begin{figure*}[htbp] 
\centering
\includegraphics[width=1.00\textwidth]{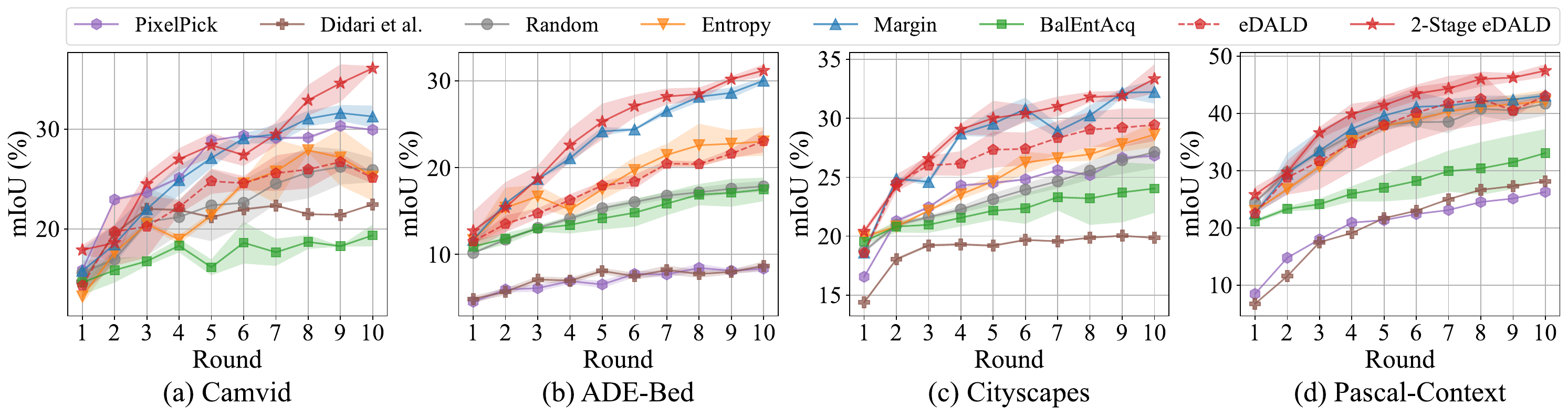}
\caption{
Active learning progression: mIoU (\%) \vs\ AL rounds on (a) CamVid, (b) ADE‐Bed, (c) Cityscapes, and (d) Pascal‐Context.
Curves compare Random sampling, single‐stage uncertainty methods, and two‐stage variants.
}
\label{fig:exp_round}
\end{figure*}

\paragraph{Round-Wise Learning Curves.}
\Cref{fig:exp_round} plots mIoU over $10$ AL rounds for various methods across four datasets.
Similar to \cref{tab:al_results}, PixelPick~\citep{pixelpick2021shin} and \citet{didari2024bayesian} use the DeepLabV3 backbone, whereas the others are based on DDPM.
Across all benchmarks, our two-stage scheme (Herding $\to$ eDALD) consistently achieves the highest performance by the final rounds.
In contrast, the single-stage uncertainty-based methods (Entropy, Margin, BalEntAcq, and eDALD) show only gradual improvements throughout the rounds, leading to comparatively smaller overall gains.
Together, these results highlight the effectiveness of the proposed two-Stage eDALD strategy.

\paragraph{Convergence to Fully‐Supervised Performance.}
\label{sec:exp:fsl}
We evaluate how quickly each method approaches fully supervised performance under a tiny per–round budget \(b=0.1N\).
The fully supervised mIoUs are: ADE-Bed: \(45.58\), CamVid: \(52.22\), Cityscapes: \(43.04\), Pascal-Context: \(60.68\).
Motivated by our low-budget findings and the adaptive-querying perspective, we use a simple two-phase schedule: 
rounds \(1\!\!-\!\!10\) (the extreme low-budget phase) use our two-stage eDALD pipeline (coverage \(\to\) uncertainty) to establish broad, non-redundant coverage; for rounds \(>10\) we drop the coverage stage and continue with uncertainty-only (Margin). 
This switch reflects two observations: (i) once coverage is established, diversity offers diminishing returns while residual errors concentrate near boundaries, where Margin is effective; and (ii) removing Stage~1 slightly reduces overhead while accelerating late-phase convergence.

\begin{figure*}[htbp]
\centering
\includegraphics[width=\textwidth]{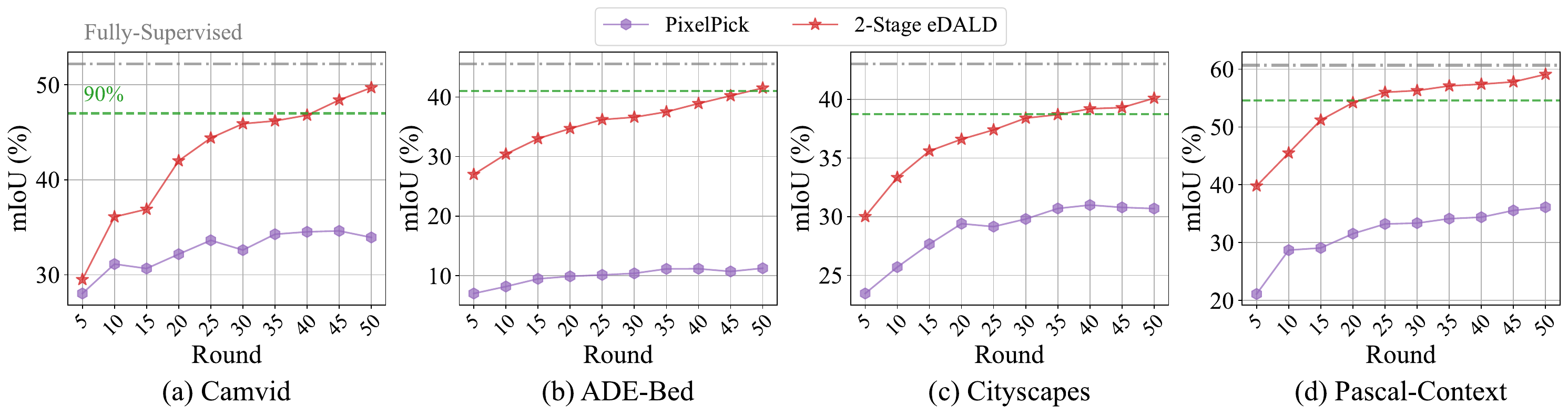}
\caption{
mIoU \vs\ AL rounds (budget $b=0.1N$).
Horizontal gray lines denote fully supervised mIoUs, and green lines denote the \(90\%\) thresholds.
We use two-stage eDALD for rounds \(1\!-\!10\) and switch to Margin thereafter.
}
\label{fig:rounds_to_sup_curves}
\end{figure*}

\Cref{fig:rounds_to_sup_curves} shows mIoU \vs\ AL rounds at \(b=0.1N\).
Across all four datasets, the proposed schedule consistently reaches \(90\%\) of the fully supervised mIoU in far fewer rounds than PixelPick~\cite{pixelpick2021shin}.
Concretely, two-stage eDALD (early) \(\to\) Margin (late) attains the \(90\%\) target within \(21\!\!-\!\!47\) rounds -- specifically, CamVid: \(32\) rounds, ADE-Bed: \(47\), Cityscapes: \(28\), Pascal-Context: \(21\) -- while labeling only \(0.003\%\!-\!0.007\%\) of all pixels in total.

\subsection{Qualitative Analysis}
\label{sec:qual}

\begin{figure*}[htbp]
\centering
\includegraphics[width=\textwidth]{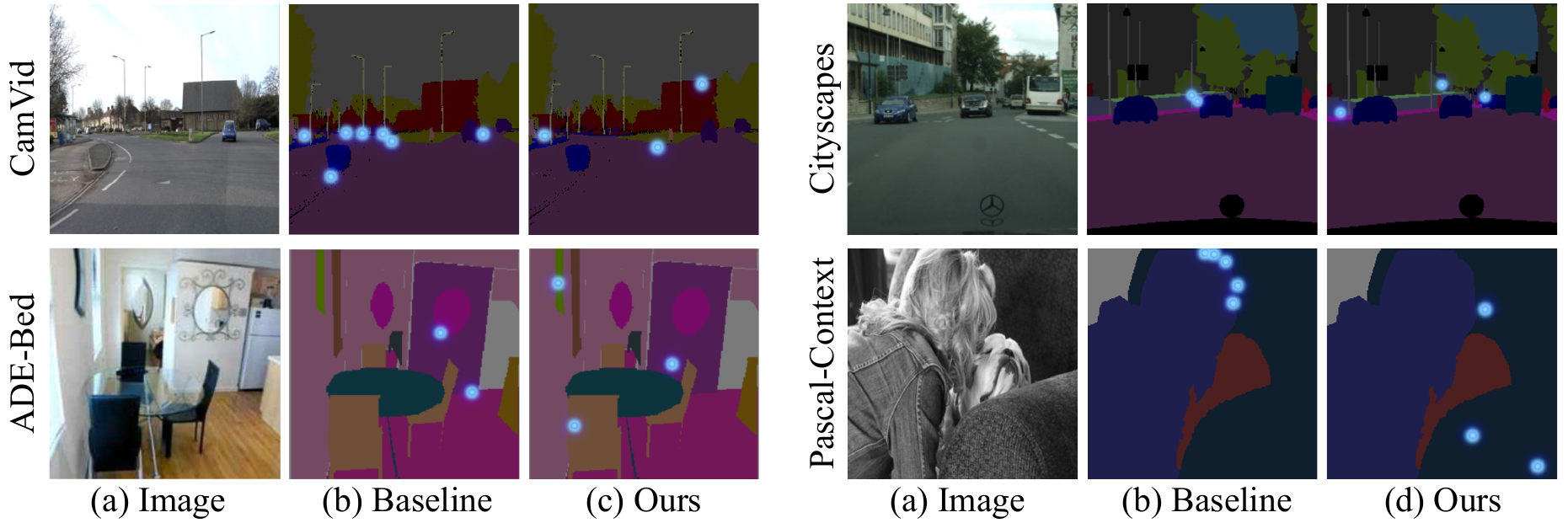}
\caption{
Qualitative pixel‐selection comparison. (a) Input image; (b) baseline selections; (c) our selections. Light blue dots indicate selected pixels. Our method covers boundaries and fine details more broadly.
}
\label{fig:qual}
\end{figure*}

\Cref{fig:qual} compares pixel selections on four examples from different benchmarks.
Margin (Baseline) tends to select overlapping pixels mostly near object boundaries whereas our two‐stage method evenly covers object boundaries, thin structures, and small classes -- even under extremely low pixel budgets.

\section{Conclusion}
\label{sec:conclusion}

We addressed the problem of \emph{low-budget active learning for semantic segmentation} by introducing a novel, two-stage pipeline that first narrows down candidate pixels through \emph{clustering with diffusion-based representations}, then refines the selection via \emph{disagreement-based uncertainty sampling}. 
Our approach effectively balances diversity and informativeness under extreme labeling constraints, as validated on four standard benchmarks.
This work would open up a new direction for integrating advanced generative backbones such as diffusion models, to budget-constrained annotation pipelines, and paves the way for more practical, large-scale semantic segmentation systems.

\paragraph{Limitations.} The proposed diffusion-based active learning by disagreement (DALD) is tailored for a diffusion-backbone; it requires sampling based on different random noises. Thus, DALD is applicable only with the diffusion-backbone. However, entropy-augmentation and two-stage sampling based on MaxHerding is agnostic to the choice of backbones. 

\clearpage

\begin{ack}
This work was supported by Institute of Information \& communications Technology Planning \& Evaluation (IITP) grant (No. RS-2022-00155966, Artificial Intelligence Convergence Innovation Human Resources Development (Ewha Womans University)) and the National Research Foundation of Korea (NRF) grant (No. RS-2025-16070597) funded by the Korea government (MSIT),
as well as by the Natural Sciences and Engineering Research Council of Canada,
the Canada CIFAR AI Chairs program,
Calcul Québec,
the BCI DRI Group,
and the Digital Research Alliance of Canada.
\end{ack}

\printbibliography

\clearpage
\appendix

\phantomsection
\begin{center}
{\LARGE\bfseries Appendix}
\end{center}

\addcontentsline{toc}{section}{Appendix}

\section{Broader Impacts}  

Our two‐stage low‐budget active learning pipeline can reduce annotation requirements by orders of magnitude, making high‐quality semantic segmentation accessible in domains with scarce labeling resources -- such as medical imaging (\eg histopathology, radiology), environmental monitoring (\eg land‐cover mapping, wildlife surveys), and infrastructure inspection in developing regions.
By building on a frozen, pre‐trained diffusion backbone, we also cut down on repeated large‐scale training, thereby lowering computational demands and carbon emissions.
This democratizes cutting‐edge segmentation tools for academic labs, start-ups, and NGOs that lack extensive annotation budgets or compute clusters.

On the other hand, any automated selection mechanism risks perpetuating biases present in the pre‐trained diffusion model or in the small initial labeled set -- potentially under‐representing rare or sensitive classes (\eg skin lesions in medical scans, minority populations in urban scenes). Moreover, ease of pixel‐level segmentation could be misused for large‐scale surveillance or privacy‐invasive monitoring if deployed without strict governance. We therefore recommend rigorous bias audits, transparent reporting of model behavior on under‐represented groups, and adherence to data‐privacy regulations when applying our approach in sensitive settings.  

\section{Implementation Details}
\label{appendix:hparams}

This section provides all settings needed to reproduce our experiments.
All code, configuration files, and training/evaluation scripts are publicly available at
\url{https://github.com/jn-kim/two-stage-edald}.

\subsection{Hyperparameters}
\label{sec:hyperparams}

Table~\ref{tab:hparams} summarizes the key hyperparameters and implementation details used across the datasets in our experiments.

\newcolumntype{D}{>{\centering\arraybackslash}p{6.3em}}
\begin{table}[htbp]
\centering
\setlength{\tabcolsep}{8pt}
\caption{Hyperparameters and implementation details across datasets.}
\begin{tabular}{c|DDDD}
\toprule
\textbf{Details} & \textbf{ADE-Bed} & \textbf{CamVid} & \textbf{Cityscapes} & \textbf{Pascal VOC} \\
\midrule
Diffusion model & \texttt{lsun\_bedroom} & \texttt{imagenet\_256} & \texttt{imagenet\_256} & \texttt{imagenet\_256} \\
Resize method & Center crop & Bilinear & Center crop & Bilinear \\
Image resolution & \multicolumn{4}{c}{$256\times256$} \\
Diffusion Steps & \multicolumn{4}{c}{$\{50, 150, 250\}$} \\
Feature blocks & \multicolumn{4}{c}{$\{5, 8, 12, 17\}$} \\
Batch size & \multicolumn{4}{c}{$5$} \\
Learning rate & \multicolumn{4}{c}{$1\times10^{-3}$} \\
Weight decay & \multicolumn{4}{c}{$1\times10^{-5}$} \\
Optimizer & \multicolumn{4}{c}{Adam} \\
Scheduler & \multicolumn{4}{c}{cosineAnnealingLR} \\
Classifier & \multicolumn{4}{c}{MLP} \\
\bottomrule
\end{tabular}%
\label{tab:hparams}
\end{table}

\noindent\textbf{Notes:}  
\begin{itemize}[leftmargin=*]
    \item \texttt{imagenet\_256} refers to an unconditional ImageNet diffusion model trained at 256$\times$256 resolution.
    \item \texttt{lsun\_bedroom} refers to a diffusion model trained on LSUN bedroom data (three classes) at 256$\times$256 resolution.
    \item Diffusion steps and feature blocks were chosen based on preliminary analysis to capture both low- and high-level semantics; see Section~\ref{sec:exp:block} for further discussion.
\end{itemize}


\subsection{Feature Extraction from the Diffusion Model}
\label{sec:feature_extraction}

To build comprehensive multi-scale representations, we extract intermediate features from a pre-trained diffusion model at denoising steps $t \in \{50, 150, 250\}$. These timesteps were selected from the later steps of the reverse diffusion process, as prior studies~\cite{baranchuk2022ledm} have shown that activations at these stages yield more discriminative semantic features, thereby enhancing pixel-level predictions.
For each timestep, we collect feature maps from selected decoder blocks (specifically, blocks $5$, $8$, $12$, and $17$). 
The exact shapes of the outputs of these blocks (excluding skip connections) during a forward pass are summarized in Table~\ref{tab:feature_shapes}. 
These feature maps are first resized (using bilinear interpolation) to the input resolution ($256\times256$), and then concatenated channel-wise to form the final pixel-level representation.

\begin{table}[htbp]
\centering
\small
\setlength{\tabcolsep}{16pt}
\caption{Output shapes of feature blocks extracted from the diffusion model. Highlighted rows indicate the selected blocks.}
\label{tab:feature_shapes}
\begin{tabular}{cc|cc}
\toprule
\textbf{Block Index} & \textbf{Output Shape} & \textbf{Block Index} & \textbf{Output Shape} \\
\midrule
Block 0  & [1024, 8, 8]   & Block 9  & [512, 64, 64] \\
Block 1  & [1024, 8, 8]   & Block 10 & [512, 64, 64]  \\
Block 2  & [1024, 16, 16] & Block 11 & [512, 128, 128] \\
Block 3  & [1024, 16, 16] & \cellcolor{gray!20}\textbf{Block 12} & \cellcolor{gray!20}\textbf{[256, 128, 128]} \\
Block 4  & [1024, 16, 16] & Block 13 & [256, 128, 128] \\
\cellcolor{gray!20}\textbf{Block 5}  & \cellcolor{gray!20}\textbf{[1024, 32, 32]} & Block 14 & [256, 256, 256] \\
Block 6  & [512, 32, 32]  & Block 15 & [256, 256, 256] \\
Block 7  & [512, 32, 32]  & Block 16 & [256, 256, 256] \\
\cellcolor{gray!20}\textbf{Block 8}  & \cellcolor{gray!20}\textbf{[512, 64, 64]}  & \cellcolor{gray!20}\textbf{Block 17} & \cellcolor{gray!20}\textbf{[256, 256, 256]} \\
\bottomrule
\end{tabular}
\end{table}

\paragraph{Concatenation Strategy.}  
For each pixel, given $T=3$ timesteps and $L=4$ selected blocks per timestep, we resize and concatenate the corresponding feature maps to obtain a feature of shape:
\[
\Bigl(T \times \bigl[\text{channels from Block 5} + \text{Block 8} + \text{Block 12} + \text{Block 17}\bigr],\, 256,\, 256\Bigr).
\]
For instance, if Block 5 outputs 1024 channels, Block 8 outputs 512 channels, and both Blocks 12 and 17 output 256 channels each, the final representation will have 
\[
3 \times (1024+512+256+256) = 3 \times 2048 = 6144 \quad \text{channels.}
\]

\subsection{Structure and Training of the Segmentation Head}

For pixel classification, we use a multilayer perceptron (MLP) with two hidden layers, incorporating ReLU activations and batch normalization layers, following the architecture used in~\cite{baranchuk2022ledm}. The classifier takes a feature vector for each pixel as input and outputs a softmax probability distribution over the classes.
Its parameters are updated using Adam optimizer with an initial learning rate of $1\times10^{-3}$ and a weight decay of $1\times10^{-5}$.
We apply a cosine annealing learning rate scheduler with $\ensuremath{T_{\text{max}}}=5$ and $\ensuremath{\eta_{\text{min}}}=1\times10^{-6}$.


\section{Ablation Studies}
\label{sec:ablation}




\subsection{Hyperparameter Sensitivity}
\label{sec:exp:hyperparameter}

We study how our two‐stage pipeline responds to the key hyperparameters \(K\) (per‐image representatives) and \(M\) (global pool size) on CamVid.  We vary 
\[
K \in \{20,\,50,\,100\}, 
\quad M \in \{0.25M_0,\,0.4M_0,\,0.5M_0\}, 
\]
where \(M_0 = N \times K\) is the size of the merged local pool (or initial global pool).  

\begin{figure*}[t!]
\centering
\includegraphics[width=0.85\textwidth]{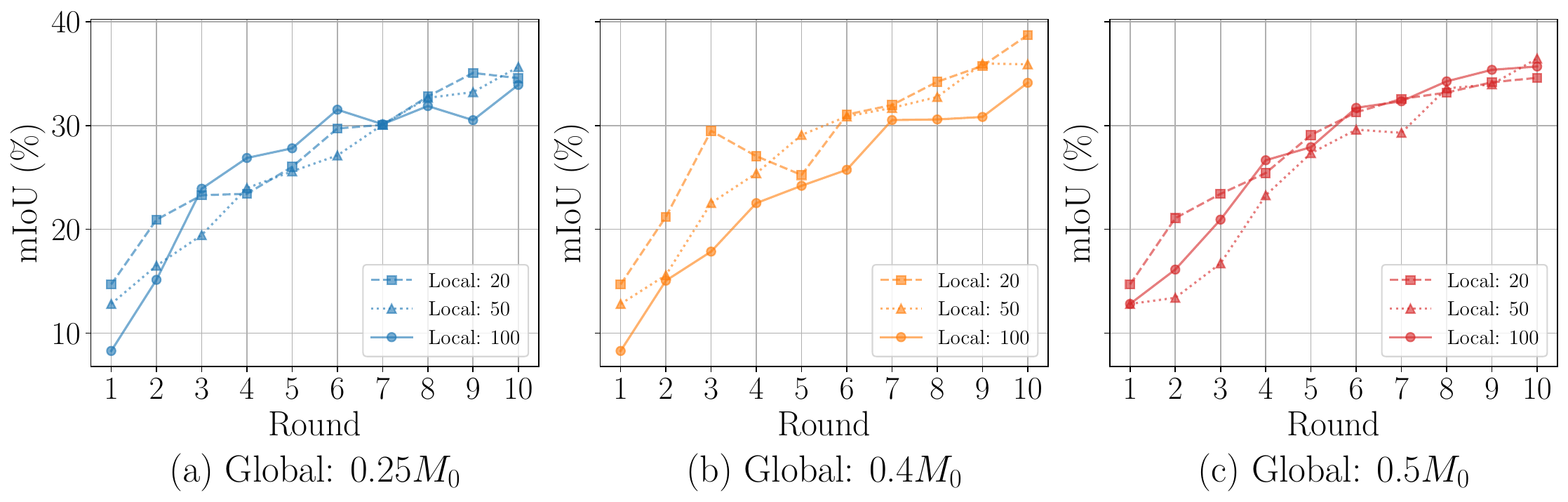}
\caption{
Hyperparameter sensitivity on CamVid: mIoU achieved after 10 AL rounds for varying per-image representative count \(K\in\{20,50,100\}\) and global pool size \(M\in\{0.25M_0,0.4M_0,0.5M_0\}\).
Each subplot shows one \(M\) setting.
}
\label{fig:hyperparams}
\end{figure*}

\Cref{fig:hyperparams} shows that our default hyperparameter \((K{=}100,\,M{=}0.5M_0)\) yields $36.12$ mIoU, while the best configuration \((K{=}20,\,M{=}0.4M_0)\) achieves $38.70$ mIoU.
This indicates that reducing \(K\) may not degrade -- and can even improve -- performance when paired with a suitably scaled global pool.
Section~\ref{sec:exp:complexity} discusses the resulting compute–accuracy trade‐offs in more detail.

\subsection{Budget Sensitivity}
\label{sec:exp:budget}

\begin{figure*}[t!]
\centering
\includegraphics[width=\textwidth]{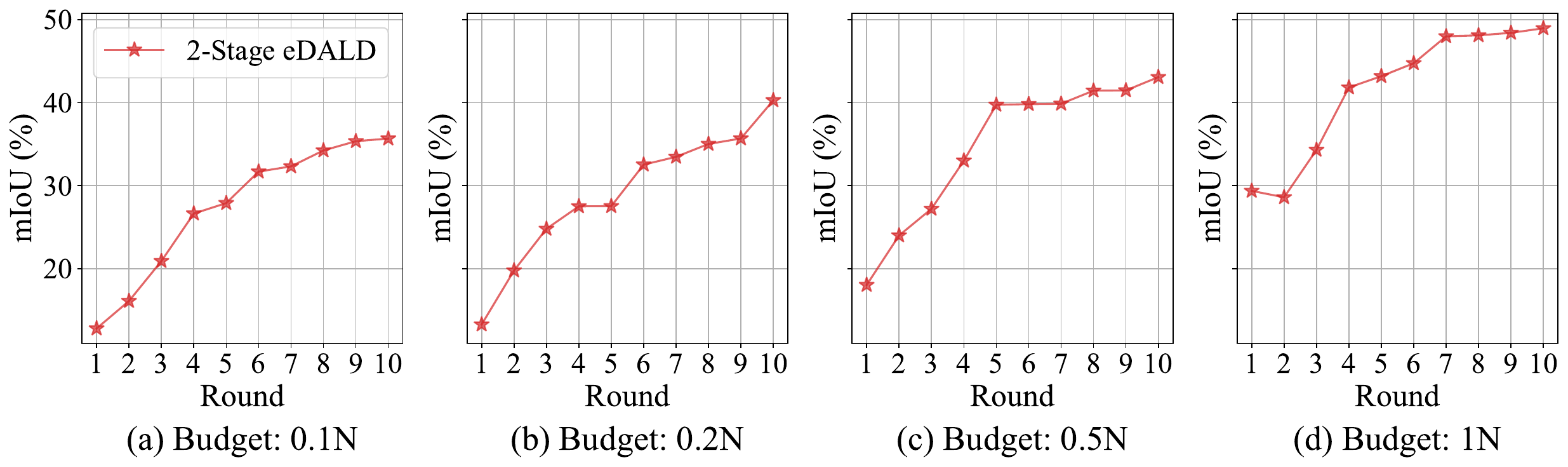}
\caption{
Results on CamVid with various budgets: \(0.1N\), \(0.2N\), \(0.5N\), and \(1N\) pixels per round.
}
\label{fig:budgets}
\end{figure*}

\Cref{fig:budgets} plots the mIoU of our two‐stage eDALD over $10$ AL rounds on CamVid under four annotation budgets: \(0.1N\), \(0.2N\), \(0.5N\), and \(1N\) pixels per round (with \(0.1N\approx0.0015\%\) of all pixels).
Under the smallest budget ($0.1N$), two-stage eDALD converges to $\sim\!35.7\%$ mIoU but improves more rapidly in early rounds.
As the budget increases, the final (round $10$) mIoU of eDALD rises accordingly: $40.3\%$ at $0.2N$, $43.1\%$ at $0.5N$, and $48.9\%$ at $1N$.  


\begin{table}[b!]
\centering
\caption{
Rounds and labeled-pixel fraction (percentage of all pixels) needed to reach 80\% and 90\% of the fully supervised mIoU (FSL) under the \(b=0.1N\) budget. 
We use two-stage eDALD for rounds \(1\!-\!10\) and switch to Margin thereafter.
}
\vspace{3pt}
\label{tab:rounds_to_thresholds}
\small
\setlength{\tabcolsep}{10pt}
\begin{tabular}{c|c|cccc}
\toprule
\textbf{Target} & \textbf{Metric} & \textbf{CamVid} & \textbf{ADE‐Bed} & \textbf{Cityscapes} & \textbf{Pascal‐C} \\
\midrule
\multirow{2}{*}{80\% FSL} 
  & Rounds    & 19 & 27 & 12 & 11 \\
  & \%Pixels  & 0.0031\% & 0.0043\% & 0.0020\% & 0.0018\% \\
\midrule
\multirow{2}{*}{90\% FSL} 
  & Rounds    & 32 & 47 & 28 & 21 \\
  & \%Pixels  & 0.0050\% & 0.0073\% & 0.0044\% & 0.0034\% \\
\bottomrule
\end{tabular}
\end{table}

\subsection{Convergence to Fully Supervised: Additional Details}
This section expands on the convergence analysis discussed in Section~\ref{sec:exp:fsl}. 
Table~\ref{tab:rounds_to_thresholds} reports, for each dataset, the number of rounds and labeled-pixel fractions required to reach \(80\%\) and \(90\%\) of the fully supervised mIoU under budget \(b=0.1N\) (\ie on average \(0.1\) pixel/image/round).
We use two-stage eDALD for rounds \(1\!-\!10\) and switch to Margin thereafter.

Across datasets, two-stage eDALD (rounds 1--10) followed by Margin converges to strong performance with \emph{vanishingly few} labeled pixels.
Reaching $80\%$ of fully supervised mIoU takes only 11--27 rounds, corresponding to 0.0018--0.0043\% of all pixels.
For $90\%$, convergence occurs within 21--47 rounds, using merely 0.0034--0.0073\% of pixels (\(\approx\) one out of \(30\text{k}\) to \(14\text{k}\) pixels).
Dataset-wise, Pascal-Context is fastest to 90\% (21 rounds, 0.0034\%), followed by Cityscapes (28, 0.0044\%) and CamVid (32, 0.0050\%), while ADE-Bed is most demanding (47, 0.0073\%), likely reflecting indoor-scene variability and pretraining mismatch.
These results substantiate our switching policy: diversity $\to$ uncertainty (two-stage) is most sample-efficient under extreme budgets, and a late switch to pure uncertainty (Margin) accelerates the final approach to the supervised ceiling.
Overall, near-supervised quality is attainable at labeling rates that are orders of magnitude below \(1\%\), reinforcing the practicality of our low-budget AL setting.

\subsection{Additional Backbone Results}
\label{sec:exp:backbone}

To compare our DDPM-based two-stage eDALD with other architectures,
we extend the two-stage framework to two widely used backbones, DeepLabV3~\cite{chen2017deeplabv3} and ViT~\cite{vit2020dosovitskiy}.
Since eDALD leverages the intrinsic stochasticity of diffusion models,
we employ its counterpart, eBALD, which estimates entropy-augmented mutual information via Monte Carlo dropout for non-diffusion backbones.
The results in \Cref{tab:backbone_results} demonstrate that the DDPM-based approach consistently outperforms the others across all datasets,
highlighting the strong synergy between diffusion-driven uncertainty estimation and MaxHerding under the low-budget regime.

We further present a ViT-specific comparison in \Cref{tab:vit_comparison},
comparing the two-stage eBALD with single-stage Margin and BalEntAcq.
The two-stage approach consistently achieves the highest final-round mIoU,
demonstrating that entropy-augmented disagreement remains beneficial even for transformer-based backbones.

\begin{table}[htbp]
\centering
\small
\caption{mIoU (\%) of two-stage methods under a low-budget regime (10 rounds) with three backbones: DeepLabV3, ViT, and DDPM. 
DDPM uses 2-Stage eDALD, while DeepLabV3 and ViT use 2-Stage eBALD.}
\setlength{\tabcolsep}{10pt}
\begin{tabular}{l|cccc}
\toprule
\textbf{Backbone} & \textbf{CamVid} & \textbf{ADE-Bed} & \textbf{Cityscapes} & \textbf{Pascal-C} \\
\midrule
DeepLabV3~\cite{chen2017deeplabv3} & 29.48 & 9.75 & 31.64 & 28.52 \\
ViT~\cite{vit2020dosovitskiy}    & 31.48 & 10.80 & 32.27 & 31.98 \\
DDPM~\cite{ho2020denoising}      & \textbf{36.12} & \textbf{31.12} & \textbf{33.34} & \textbf{47.98} \\
\bottomrule
\end{tabular}
\label{tab:backbone_results}
\end{table}

\begin{table}[htbp]
\centering
\small
\caption{ViT-specific comparison of active learning methods in mIoU (\%) under the low-budget regime.}
\setlength{\tabcolsep}{10pt}
\begin{tabular}{l|cccc}
\toprule
\textbf{Method} & \textbf{CamVid} & \textbf{ADE-Bed} & \textbf{Cityscapes} & \textbf{Pascal-C} \\
\midrule
Margin        & 29.52 & 9.51 & 27.36 & 29.43 \\
BalEntAcq     & 24.67 & 9.81 & 21.05 & 30.53 \\
2-Stage eBALD & \textbf{31.48} & \textbf{10.80} & \textbf{32.27} & \textbf{31.98} \\
\bottomrule
\end{tabular}
\label{tab:vit_comparison}
\end{table}

\subsection{Comparison with Diversity-based Methods}
\label{sec:exp:coreset}

To assess the effectiveness of different diversity-driven selection strategies,
we compare pipelines using either MaxHerding~\cite{bae2024maxherding} or Core-set~\cite{sener2017active} as the first-stage selector,
followed by eDALD for uncertainty-based refinement.
Core-set corresponds to the $k$-center greedy algorithm, which iteratively selects the farthest sample in feature space to ensure coverage.
Table~\ref{tab:diversity_comparison} summarizes the final-round mIoU across four datasets,
comparing both one-stage and two-stage variants with two backbones: DeepLabV3 and DDPM.

\begin{table*}[htbp]
\centering
\caption{Final-round mIoU comparison of diversity methods (Core-set \vs\ MaxHerding) under the low-budget regime.  
For DeepLab we report two-stage variants with eBALD, while for DDPM we report both one-stage and two-stage with eDALD.}
\small
\setlength{\tabcolsep}{6pt}
\begin{tabular}{>{\centering\arraybackslash}m{2cm}|l|
>{\centering\arraybackslash}m{1.5cm}%
>{\centering\arraybackslash}m{1.5cm}%
>{\centering\arraybackslash}m{1.5cm}%
>{\centering\arraybackslash}m{1.5cm}}
\toprule
\textbf{Backbone} & \textbf{Method} & \textbf{ADE-Bed} & \textbf{CamVid} & \textbf{Cityscapes} & \textbf{Pascal-C} \\
\midrule
\multirow{4}{*}{\centering DeepLabV3~\cite{chen2017deeplabv3}} 
  & Core-set & 9.21 & 26.49 & 30.42 & 25.83 \\
  & MaxHerding & 9.64 & 28.78 & 31.05 & 28.12 \\
  & Core-set $\rightarrow$ eBALD & 9.53 & 26.50 & 30.48 & 26.70 \\
  & MaxHerding $\rightarrow$ eBALD & 9.70 & 29.24 & 31.45 & 28.50 \\
\midrule
\multirow{4}{*}{\centering DDPM~\cite{ho2020denoising}} 
  & Core-set & 18.49 & 16.19 & 22.74 & 23.27 \\
  & MaxHerding & 24.06 & 31.83 & 25.80 & \textbf{52.04} \\
  & Core-set $\rightarrow$ eDALD & 26.70 & 32.95 & 32.20 &  44.58 \\
  & MaxHerding $\rightarrow$ eDALD & \textbf{31.12} & \textbf{36.12} & \textbf{33.34} & 47.98 \\
\bottomrule
\end{tabular}
\label{tab:diversity_comparison}
\end{table*}

The results reveal two consistent trends.
First, MaxHerding provides stronger coverage than Core-set across all configurations, whether used alone or within a two-stage pipeline.
Second, this advantage is consistent across both backbones but is particularly pronounced for DDPM,
where the combination MaxHerding $\to$ eDALD achieves the best overall performance.
An exception is observed on Pascal-Context, where MaxHerding alone achieves higher performance than its two-stage variant.
We attribute this to the dataset’s large class count ($33$) and diverse scene distribution, where maximizing pure feature-space coverage can be more effective for capturing rare or spatially scattered classes.
Overall, these findings demonstrate that while MaxHerding alone is a strong representation-based baseline,
its combination with eDALD uncertainty consistently yields superior performance in most practical scenarios.

\subsection{Pixel-to-Region Expansion with SAM}
\label{sec:exp:sam_practical}
We extend our two-stage eDALD pipeline to practical \emph{region}-level supervision by integrating SAM~\cite{kirillov2023segment} under a strict \emph{click-parity} budget (\(b{=}0.1N\), \ie on average \(0.1\) pixel/image/round).
Each selected pixel is used as \emph{one} positive point prompt for SAM (no extra clicks and no iterative refinements). 
SAM returns multiple mask proposals with predicted IoU scores; we keep the highest-scoring proposal if it exceeds a confidence threshold \(\tau\) and otherwise fall back to the single-pixel label. 
Thus, each click can expand to a dense region label when reliable, yielding many more supervised pixels without increasing annotation cost.

Table~\ref{tab:round_performance_main} reports 10-round results for the baseline two-stage eDALD (w/o SAM) and the SAM-augmented variant (w/ SAM).
Across datasets, SAM consistently \emph{accelerates early rounds} and often improves final mIoU under the same budget: on CamVid the final gain is $+5.24$\,pp ($41.34$ \vs\ $36.10$), while ADE-Bed, Cityscapes, and Pascal-Context show modest but steady improvements ($+0.48$\,pp, $+0.68$\,pp, and $+0.27$\,pp, respectively).
Overall, single-click, region-level supervision via SAM provides faster convergence and better cost-efficiency in the extreme low-budget setting.

\begin{table*}[htbp]
\centering
\caption{
Active learning performance (mIoU, \%) over 10 rounds for two-stage eDALD with and without SAM under the \(b{=}0.1N\) budget (\emph{strict click parity}: one positive point per selected pixel).}
\label{tab:round_performance_main}
\small
\setlength{\tabcolsep}{4.2pt}
\renewcommand{\arraystretch}{1}
\begin{tabular}{ll| *{10}{c}} 
\toprule
\textbf{Dataset} & \textbf{Method} & \textbf{R1} & \textbf{R2} & \textbf{R3} & \textbf{R4} & \textbf{R5} & \textbf{R6} & \textbf{R7} & \textbf{R8} & \textbf{R9} & \textbf{R10} \\
\midrule
\multirow{2}{*}{CamVid}
& w/o SAM & 17.90 & 18.63 & 24.52 & 27.00 & 28.40 & 27.40 & 29.50 & 32.90 & 34.60 & 36.10 \\
& w/ SAM & 20.51 & 28.68 & 32.03 & 28.74 & 31.93 & 34.78 & 37.75 & 36.74 & 39.56 & 41.34 \\
\midrule
\multirow{2}{*}{ADE-Bed}
& w/o SAM & 12.70 & 15.40 & 18.70 & 22.60 & 25.30 & 27.10 & 28.20 & 28.50 & 30.20 & 31.20 \\
& w/ SAM & 14.00 & 16.37 & 19.94 & 24.13 & 29.11 & 29.58 & 30.34 & 31.49 & 32.43 & 31.68 \\
\midrule
\multirow{2}{*}{Cityscapes}
& w/o SAM & 20.40 & 24.20 & 26.57 & 29.06 & 30.01 & 30.40 & 31.00 & 31.80 & 31.90 & 33.34 \\
& w/ SAM & 20.31 & 25.85 & 27.20 & 26.37 & 26.55 & 27.20 & 28.05 & 29.83 & 33.89 & 34.02 \\
\midrule
\multirow{2}{*}{Pascal-Context}
& w/o SAM & 25.80 & 29.75 & 36.61 & 39.89 & 41.42 & 43.43 & 44.33 & 45.98 & 46.24 & 47.44 \\
& w/ SAM & 30.42 & 38.98 & 40.45 & 41.90 & 42.33 & 43.22 & 43.48 & 44.53 & 46.87 & 47.71 \\
\bottomrule
\end{tabular}
\end{table*}



\subsection{Decoder‐Block Selection Ablation}
\label{sec:exp:block}

We compare four decoder‐block configurations for multi‐scale feature extraction on ADE‐Bed:
\begin{itemize}[leftmargin=*]
  \item \(\{2,3\}\): deep blocks (high-dim features, low spatial resolution)
  \item \(\{11\text{--}17\}\): shallow blocks (low‐dim features, high spatial resolution)
  \item \(\{5,6,7,8,12\}\): original LEDM setting~\citep{baranchuk2022ledm}
  \item \(\{5,8,12,17\}\): our compact selection
\end{itemize}

Table~\ref{tab:block_ablation} reports mIoU for each.
The compact set \(\{5,8,12,17\}\) ($45.58\%$) matches the original five‐block configuration ($46.41\%$), while deep-only \(\{2,3\}\) ($42.19\%$) and shallow‐only \(\{11\text{--}17\}\) ($22.68\%$) both degrade significantly. 
This confirms that retaining both spatial detail and high‐level semantics is crucial.
We therefore adopt \(\{5,8,12,17\}\) in all experiments -- it reduces feature dimensionality by $\sim\!28\%$ and speeds up without sacrificing accuracy.

\begin{table}[htbp]
\centering
\caption{
Ablation on decoder‐block choice (ADE‐Bed). Channel dim.\ is the sum over selected blocks.
}
\label{tab:block_ablation}
\vspace{2mm}
\small
\setlength{\tabcolsep}{10pt}
\begin{tabular}{lc|c}
\toprule
\textbf{Blocks}            & \textbf{Channel Dim.} & \textbf{mIoU (\%)} \\
\midrule
\(\{2,3\}\) (deep‐only)    & 6,144                 & 42.19              \\
\(\{11\text{--}17\}\) (shallow‐only) & 6,144                 & 22.68              \\
\(\{5,6,7,8,12\}\) (original)       & 8,448                 & 46.41              \\
\(\{5,8,12,17\}\) (compact)        & 6,144                 & 45.58              \\
\bottomrule
\end{tabular}
\end{table}

\subsection{Computational Complexity}
\label{sec:exp:complexity}

The dominant cost in our pipeline is candidate selection via MaxHerding.
Given a global pool of size \(M\) with \(d\)-dimensional features and per–round budget \(B\), forming the full RBF kernel costs \(\mathcal{O}(M^{2}d)\) time and \(\mathcal{O}(M^{2})\) memory.
The subsequent greedy selection over \(B\) points adds \(\mathcal{O}(M^{2}B)\) time, giving an overall complexity of \(\mathcal{O}\big(M^{2}\!\cdot\!\max\{d,B\}\big)\).
Since \(M=N\times K\) grows with the number of images \(N\) and the per–image candidate count \(K\), this step can become the bottleneck on large datasets.

\textbf{Runtime/memory in practice.}
On ADE-Bed, our two-stage eDALD with \(N{=}964\), \(K{=}50\) (\(M{=}48{,}200\)) required:
feature extraction ($8\mathrm{m}\,22\mathrm{s}$, $25.4$\,GB), local herding ($4\mathrm{m}\,24\mathrm{s}$, $3.7$\,GB), and global herding ($14\mathrm{m}\,27\mathrm{s}$, $38.1$\,GB), compared to training ($19\mathrm{m}\,14\mathrm{s}$, $22.7$\,GB).
Overall, MaxHerding accounts for \(\sim\!41\%\) of the per-round wall-clock time, and global herding dominates peak memory ($38.1$\,GB).

\textbf{Reducing peak memory.}
To further lower the memory footprint, we adopt a split-and-herd variant: partition the global pool into memory-safe splits, allocate a proportional sub-budget to each split, and run greedy herding sequentially while conditioning on selections from earlier splits to preserve cross-split coverage.
This relaxes full optimality but prevents out-of-memory under higher \(K\) or high-resolution settings, effectively reducing \emph{peak} memory from \(\mathcal{O}(M^{2})\) to roughly \(\mathcal{O}\!\big((M/S)^{2}\big)\) per split when using \(S\) splits, with negligible impact on accuracy in our experiments.

\section{Qualitative Results}
\label{sec:qual_supp}

We qualitatively compare some active learning selection results from the proposed two‐stage eDALD against the single‐stage Margin baseline on each of four datasets.

\subsection{Final Pixel Selections}  
\begin{figure*}[htbp]
  \centering
  \includegraphics[width=\textwidth]{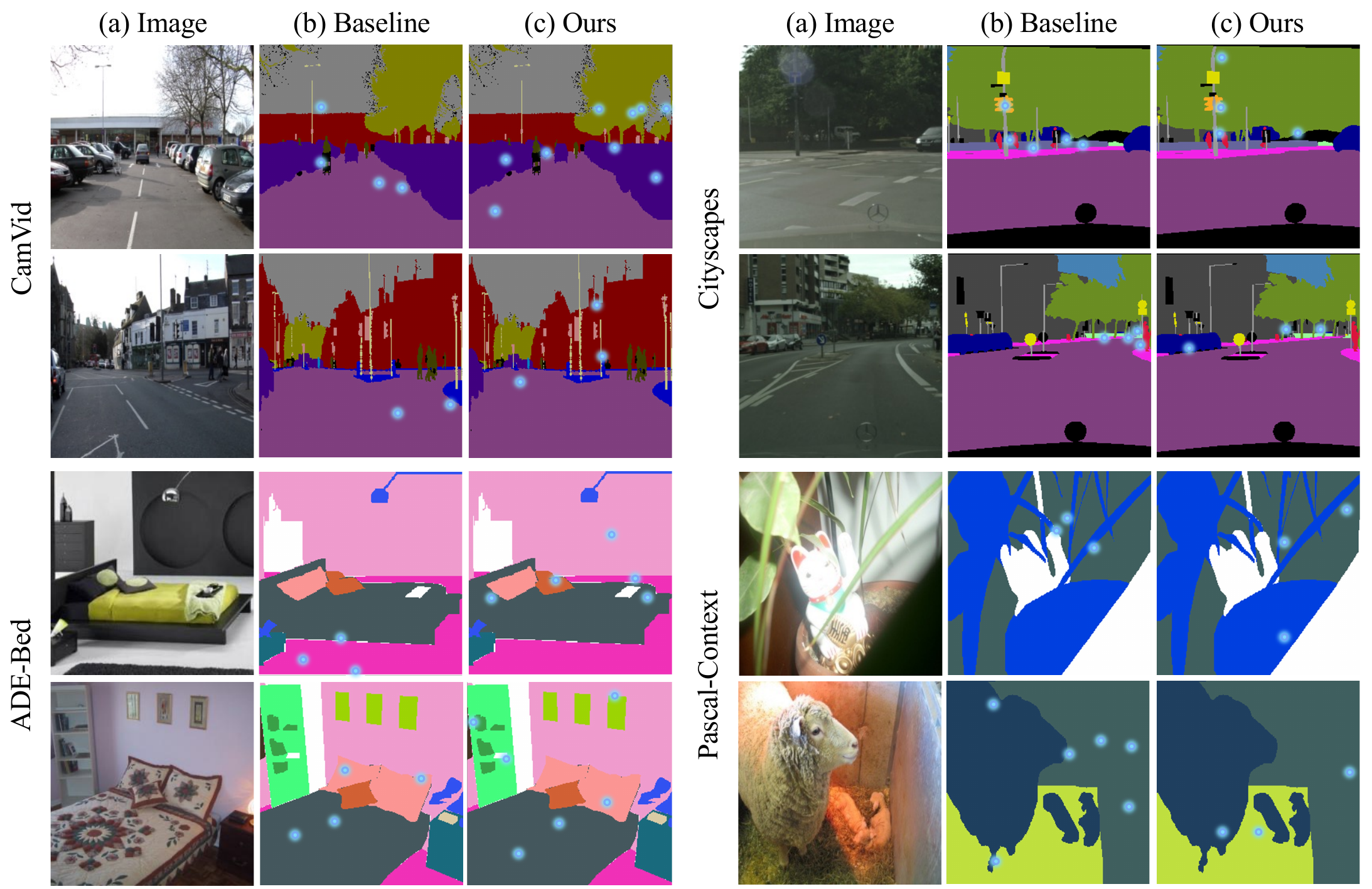}
  \caption{Final pixel selections overlaid on ground truth: (a) input image; (b) Margin selections; (c) two‐stage eDALD selections. Light blue dots mark chosen pixels.}
  \label{fig:qual_pixels}
\end{figure*}

Figure~\ref{fig:qual_pixels} overlays the \(b\) selected pixels on the ground‐truth maps for both methods.
Margin tends to concentrate selections on boundary regions and often chooses redundant pixels within the same class region, resulting in less diverse annotations.
In contrast, our two‐stage pipeline disperses pixels evenly across object edges, small classes, and scene context, ensuring that the most informative locations are annotated.

\subsection{Selection Progression}  
\begin{figure*}[htbp]
  \centering
  \includegraphics[width=\textwidth]{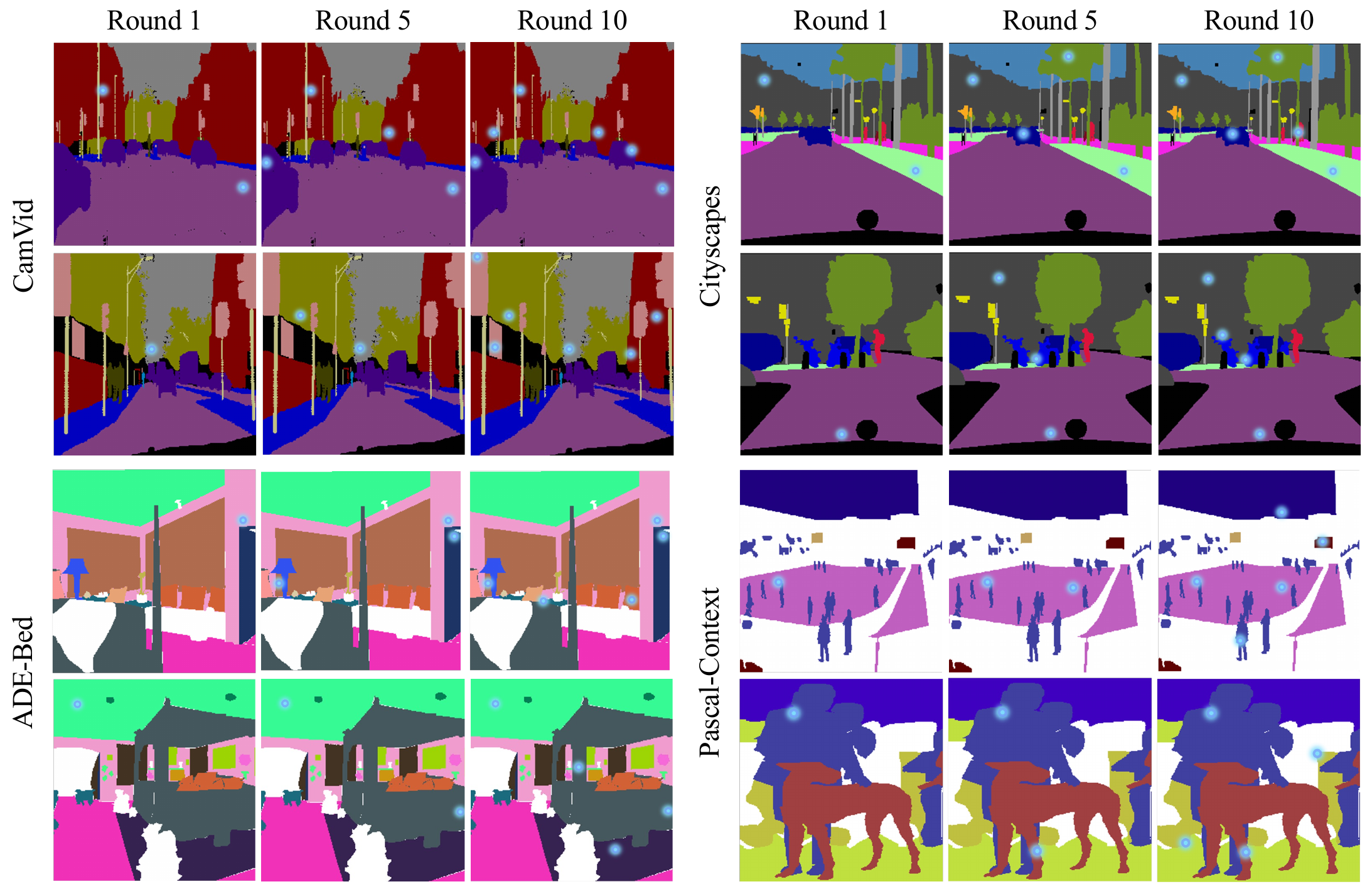}
  \caption{Two‐stage eDALD pixel selections at rounds $1$, $5$, and $10$ on multiple example images.}
  \label{fig:qual_rounds}
\end{figure*}

Figure~\ref{fig:qual_rounds} visualizes the selected pixels of two-stage eDALD across rounds.
Early rounds focus on broad structural cues; later rounds refine boundary regions and rare classes.
This progression highlights how representation filtering first ensures coverage, then eDALD uncertainty hones in on the remaining ambiguous pixels.

\subsection{Segmentation Outputs}  
\begin{figure*}[htbp]
  \centering
  \includegraphics[width=0.94\textwidth]{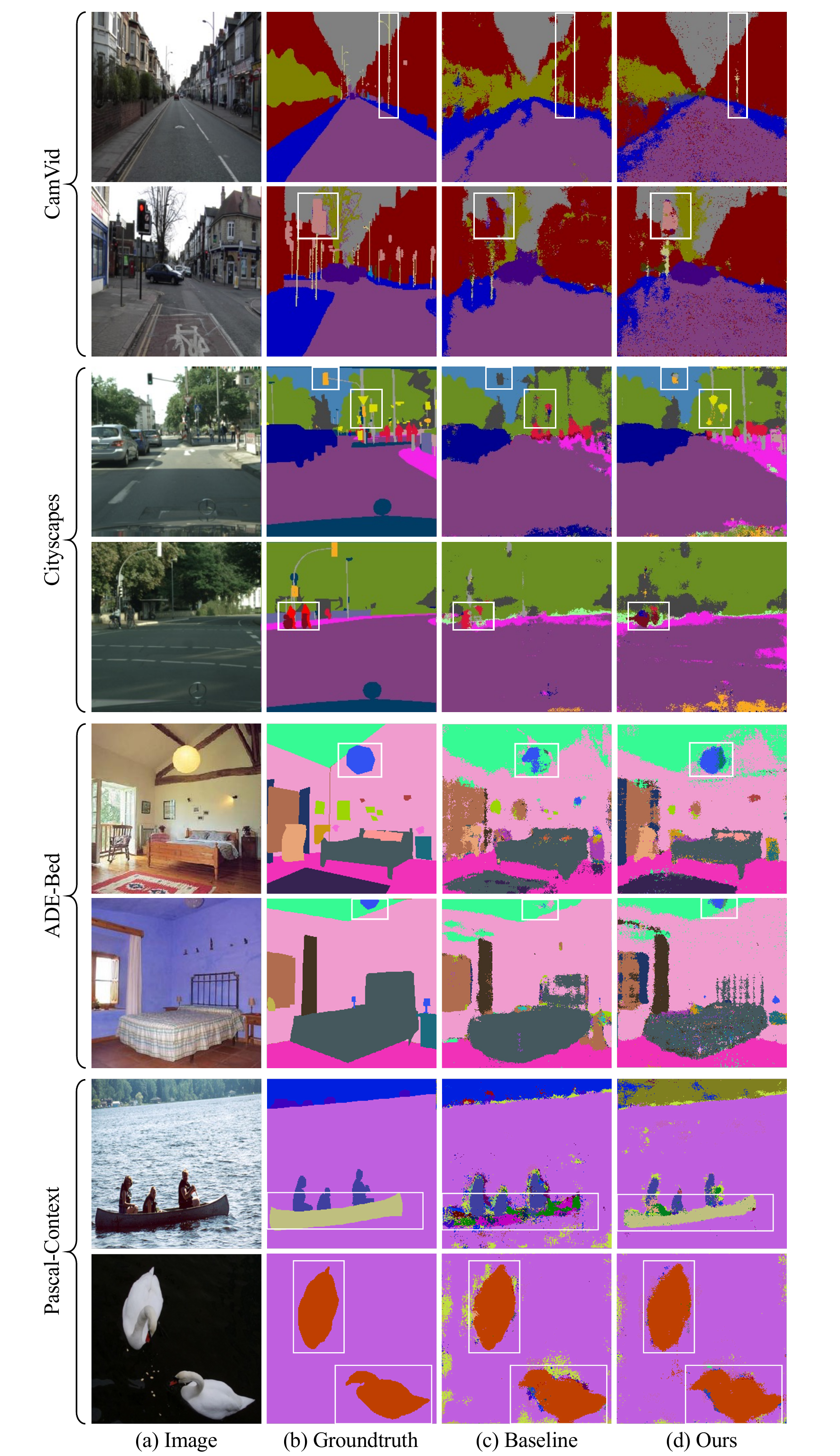}
  \caption{Example segmentation results. Columns show (a) input image, (b) ground truth, (c) Margin prediction, and (d) 2-stage eDALD prediction.}
  \label{fig:qual_seg}
\end{figure*}

For representative images, Figure~\ref{fig:qual_seg} shows the predicted segmentation maps. 
Margin often misses thin structures and small objects (\eg distant pedestrians, traffic signs), producing fragmented or smoothed regions.
In contrast, our eDALD‐trained model yields cleaner boundaries and recovers fine details more faithfully.

\end{document}

%% file: mymacros.tex
\usepackage{color}
\usepackage{algorithm}
\usepackage{algpseudocode}
\usepackage{amsmath,amssymb,amsfonts}
\usepackage{booktabs}       
\usepackage{multirow, varwidth}
\usepackage{mathtools}
\usepackage{epsfig}
\usepackage{verbatim}
\usepackage{xr}
\usepackage{flushend}
\usepackage{subcaption}
\usepackage{graphicx}
\usepackage{url}
\usepackage{enumitem}
\setlist{nosep} 
\usepackage{xcolor}         
\usepackage{kotex}
\usepackage{colortbl}

\newcommand{\argmax}{\operatornamewithlimits{argmax}}
\newcommand{\argmin}{\operatornamewithlimits{argmin}}

\DeclareRobustCommand\onedot{\futurelet\@let@token\@onedot}
\def\onedot{.} 
\def\eg{\emph{e.g}\onedot, } 
\def\ie{\emph{i.e}\onedot, } 
 
 \def\vs{\emph{vs}\onedot}

\newcommand{\x}{\mathrm{x}}

\newcommand{\vecz}{\mathbf{z}}

\newcommand{\mytilde}{\raise.17ex\hbox{$\scriptstyle\mathtt{\sim}$}}
\DeclareMathOperator*{\E}{\mathbb{E}}

\newcommand{\setL}{\mathcal{L}}
\newcommand{\setU}{\mathcal{U}}
\newcommand{\setS}{\mathcal{S}}
\newcommand{\setI}{\mathcal{I}}
\newcommand{\setM}{\mathcal{M}}
\newcommand{\setR}{\mathcal{R}}
\newcommand{\setB}{\mathcal{B}}
\newcommand{\setY}{\mathcal{Y}}
\newcommand{\seghead}{s_\theta}

\newcommand{\funcI}{\mathrm{I}}
\newcommand{\funcH}{\mathrm{H}}

\definecolor{dark2green}{rgb}{0.1, 0.65, 0.3}

\newcommand{\junhyug}[1]{{\color{blue}{#1}}}

\newcommand{\danica}[1]{{\color{violet!60}{#1}}}

\newcommand{\blue}[1]{{\color{blue}{#1}}}
\newcommand{\red}[1]{{\color{red}{#1}}}

%% file: neurips_2025.bib
@String(CVPR= {IEEE Conf. Comput. Vis. Pattern Recog.})

@String(ICCV= {Int. Conf. Comput. Vis.})

@String(BMVC= {Brit. Mach. Vis. Conf.})

@String(ICASSP=	{ICASSP})

@String(ICLR = {Int. Conf. Learn. Represent.})

@String(CVPR  = {CVPR})

@String(ICCV  = {ICCV})

@String(BMVC  =	{BMVC})

@String(ICLR  = {ICLR})

@inproceedings{kasarla2019region,
  title={Region-based active learning for efficient labeling in semantic segmentation},
  author={Kasarla, Tejaswi and Nagendar, Gattigorla and Hegde, Guruprasad M and Balasubramanian, Vineeth and Jawahar, CV},
  booktitle={2019 IEEE Winter Conference on Applications of Computer Vision (WACV)}, 
  pages={1109--1117},
  year={2019},
  organization={IEEE}
}

@article{casanova2020reinforced,
title={Reinforced active learning for image segmentation},
author={Arantxa Casanova and Pedro O. Pinheiro and Negar Rostamzadeh and Christopher J. Pal},
booktitle={ICLR},
year={2020}
}

@inproceedings{ji2023ddp,
  title={Ddp: Diffusion model for dense visual prediction},
  author={Ji, Yuanfeng and Chen, Zhe and Xie, Enze and Hong, Lanqing and Liu, Xihui and Liu, Zhaoqiang and Lu, Tong and Li, Zhenguo and Luo, Ping},
  booktitle={International Conference on Computer Vision},
  pages={21741--21752},
  year={2023}
}

@inproceedings{baranchuk2022ledm,
  title={Label-Efficient Semantic Segmentation with Diffusion Models},
  author={Baranchuk, Dmitry and Voynov, Andrey and Rubachev, Ivan and Khrulkov, Valentin and Babenko, Artem},
  booktitle={ICLR},
  year={2022}
}

@inproceedings{kim2023adaptive,
  title={Adaptive superpixel for active learning in semantic segmentation},
  author={Kim, Hoyoung and Oh, Minhyeon and Hwang, Sehyun and Kwak, Suha and Ok, Jungseul},
  booktitle={International Conference on Computer Vision},
  pages={943--953},
  year={2023}
}

@inproceedings{cai2021revisiting,
    author    = {Cai, Lile and Xu, Xun and Liew, Jun Hao and Foo, Chuan Sheng},
    title     = {Revisiting Superpixels for Active Learning in Semantic Segmentation With Realistic Annotation Costs},
    booktitle = {Proceedings of the IEEE/CVF Conference on Computer Vision and Pattern Recognition (CVPR)},
    year      = {2021},
    pages     = {10988--10997}
}

@inproceedings{sinha2019variational,
  title={Variational adversarial active learning},
  author={Sinha, Samarth and Ebrahimi, Sayna and Darrell, Trevor},
  booktitle={2019 IEEE/CVF International Conference on Computer Vision (ICCV)}, 
  pages={5971-5980},
  year={2019}
}

@inproceedings{lyu2024semi,
  title={Semi-Supervised Variational Adversarial Active Learning via Learning to Rank and Agreement-Based Pseudo Labeling},
  author={Lyu, Zongyao and Beksi, William J},
  booktitle={International Conference on Pattern Recognition},
  pages={1--16},
  year={2024},
  organization={Springer}
}

@inproceedings{xie2020deal,
  title={Deal: Difficulty-aware active learning for semantic segmentation},
  author={Xie, Shuai and Feng, Zunlei and Chen, Ying and Sun, Songtao and Ma, Chao and Song, Mingli},
  booktitle={Proceedings of the Asian conference on computer vision},
  year={2020}
}

@inproceedings{pixelpick2021shin,
    author    = {Shin, Gyungin and Xie, Weidi and Albanie, Samuel},
    title     = {All You Need Are a Few Pixels: Semantic Segmentation With PixelPick},
    booktitle = {ICCVW},
    year      = {2021},
}

@article{didari2024bayesian,
  title={Bayesian Active Learning for Semantic Segmentation},
  author={Didari, Sima and Hu, Wenjun and Woo, Jae Oh and Hao, Heng and Moon, Hankyu and Min, Seungjai},
  journal={arXiv preprint arXiv:2408.01694},
  year={2024}
}

@article{ho2020denoising,
  title={Denoising diffusion probabilistic models},
  author={Ho, Jonathan and Jain, Ajay and Abbeel, Pieter},
  journal={Advances in neural information processing systems},
  volume={33},
  pages={6840--6851},
  year={2020}
}

@article{song2019generative,
  title={Generative modeling by estimating gradients of the data distribution},
  author={Song, Yang and Ermon, Stefano},
  journal={Advances in neural information processing systems},
  volume={32},
  year={2019}
}

@article{song2020improved,
  title={Improved techniques for training score-based generative models},
  author={Song, Yang and Ermon, Stefano},
  journal={Advances in neural information processing systems},
  volume={33},
  pages={12438--12448},
  year={2020}
}

@article{song2020score,
title={Score-Based Generative Modeling through Stochastic Differential Equations},
author={Yang Song and Jascha Sohl-Dickstein and Diederik P Kingma and Abhishek Kumar and Stefano Ermon and Ben Poole},
booktitle={ICLR},
year={2021},
}

@inproceedings{rombach2022high,
  title={High-resolution image synthesis with latent diffusion models},
  author={Rombach, Robin and Blattmann, Andreas and Lorenz, Dominik and Esser, Patrick and Ommer, Bj{\"o}rn},
  booktitle={Proceedings of the IEEE/CVF conference on computer vision and pattern recognition},
  pages={10684--10695},
  year={2022}
}

@inproceedings{wu2024medsegdiff,
  title={Medsegdiff: Medical image segmentation with diffusion probabilistic model},
  author={Wu, Junde and Fu, Rao and Fang, Huihui and Zhang, Yu and Yang, Yehui and Xiong, Haoyi and Liu, Huiying and Xu, Yanwu},
  booktitle={Medical Imaging with Deep Learning},
  pages={1623--1639},
  year={2024},
  organization={PMLR}
}

@inproceedings{mittal2023best,
  title={Best practices in active learning for semantic segmentation},
  author={Mittal, Sudhanshu and Niemeijer, Joshua and Sch{\"a}fer, J{\"o}rg P and Brox, Thomas},
  booktitle={DAGM German Conference on Pattern Recognition},
  pages={427--442},
  year={2023},
  organization={Springer}
}

@inproceedings{vgg2014simonyan2,
  title={Very deep convolutional networks for large-scale image recognition},
  author={Simonyan, Karen and Zisserman, Andrew},
  booktitle={ICLR},
  year={2015}
}

@inproceedings{resnet2016he,
  title={Deep residual learning for image recognition},
  author={He, Kaiming and Zhang, Xiangyu and Ren, Shaoqing and Sun, Jian},
  booktitle={CVPR},
  pages={770--778},
  year={2016}
}

@inproceedings{vit2020dosovitskiy,
  title={An Image is Worth 16x16 Words: Transformers for Image Recognition at Scale},
  author={Dosovitskiy, Alexey and Beyer, Lucas and Kolesnikov, Alexander and Weissenborn, Dirk and Zhai, Xiaohua and Unterthiner, Thomas and Dehghani, Mostafa and Minderer, Matthias and Heigold, Georg and Gelly, Sylvain and Uszkoreit, Jakob and Houlsby, Neil},
  booktitle={ICLR},
  year={2021}
}

@inproceedings{deit2021touvron,
  title={Training data-efficient image transformers \& distillation through attention},
  author={Touvron, Hugo and Cord, Matthieu and Douze, Matthijs and Massa, Francisco and Sablayrolles, Alexandre and J{\'e}gou, Herv{\'e}},
  booktitle={International Conference on Machine Learning},
  pages={10347--10357},
  year={2021},
  organization={PMLR}
}

@inproceedings{liu2021swin,
  title={Swin transformer: Hierarchical vision transformer using shifted windows},
  author={Liu, Ze and Lin, Yutong and Cao, Yue and Hu, Han and Wei, Yixuan and Zhang, Zheng and Lin, Stephen and Guo, Baining},
  booktitle={International Conference on Computer Vision},
  pages={10012--10022},
  year={2021}
}

@article{chen2017deeplabv3,
  title={Rethinking atrous convolution for semantic image segmentation},
  author={Chen, Liang-Chieh and Papandreou, George and Schroff, Florian and Adam, Hartwig},
  journal={arXiv preprint arXiv:1706.05587},
  year={2017}
}

@inproceedings{zhou2017scene,
  title={Scene parsing through ade20k dataset},
  author={Zhou, Bolei and Zhao, Hang and Puig, Xavier and Fidler, Sanja and Barriuso, Adela and Torralba, Antonio},
  booktitle={Proceedings of the IEEE conference on computer vision and pattern recognition},
  pages={633--641},
  year={2017}
}

@article{brostow2009semantic,
  title={Semantic object classes in video: A high-definition ground truth database},
  author={Brostow, Gabriel J and Fauqueur, Julien and Cipolla, Roberto},
  journal={Pattern recognition letters},
  volume={30},
  number={2},
  pages={88--97},
  year={2009},
  publisher={Elsevier}
}

@inproceedings{cordts2016cityscapes,
  title={The cityscapes dataset for semantic urban scene understanding},
  author={Cordts, Marius and Omran, Mohamed and Ramos, Sebastian and Rehfeld, Timo and Enzweiler, Markus and Benenson, Rodrigo and Franke, Uwe and Roth, Stefan and Schiele, Bernt},
  booktitle={Proceedings of the IEEE conference on computer vision and pattern recognition},
  pages={3213--3223},
  year={2016}
}

@inproceedings{entropy2014wang,
  title={A new active labeling method for deep learning},
  author={Wang, Dan and Shang, Yi},
  booktitle={2014 International joint conference on neural networks (IJCNN)},
  pages={112--119},
  year={2014},
  organization={IEEE}
}

@inproceedings{margin2001scheffer,
  title={Active hidden markov models for information extraction},
  author={Scheffer, Tobias and Decomain, Christian and Wrobel, Stefan},
  booktitle={International symposium on intelligent data analysis},
  pages={309--318},
  year={2001},
  organization={Springer}
}

@inproceedings{heterogeneous1994lewis,
  title={Heterogeneous Uncertainty Sampling for Supervised Learning},
  author={Lewis, David D and Catlett, Jason},
  booktitle={Machine learning proceedings 1994},
  pages={148--156},
  year={1994},
  publisher={Elsevier}
}

@inproceedings{sequential1994lewis,
  title={A sequential algorithm for training text classifiers},
  author={Lewis, David D. and Gale, William A.},
  publisher = {Springer-Verlag},
  booktitle = {Proceedings of the 17th Annual International ACM SIGIR Conference on Research and Development in Information Retrieval},
  pages={3--12},
  year={1994},
}

@article{woo2021balent,
  title={Active learning in bayesian neural networks with balanced entropy learning principle},
  author={Woo, Jae Oh},
  journal={arXiv preprint arXiv:2105.14559},
  year={2021}
}

@article{kirsch2021powerbald,
  title={Stochastic batch acquisition: A simple baseline for deep active learning},
  author={Kirsch, Andreas and Farquhar, Sebastian and Atighehchian, Parmida and Jesson, Andrew and Branchaud-Charron, Frederic and Gal, Yarin},
  journal={arXiv preprint arXiv:2106.12059},
  year={2021}
}

@inproceedings{al2009settles,
  title={Active Learning Literature Survey},
  author={Settles, Burr},
  year={2009},
  publisher={University of Wisconsin-Madison Department of Computer Sciences}
}

@inproceedings{egl2007settles,
  title={Multiple-instance active learning},
  author={Settles, Burr and Craven, Mark and Ray, Soumya},
  journal={Advances in neural information processing systems},
  volume={20},
  year={2007}
}

@article{badge2019ash,
  title={Deep batch active learning by diverse, uncertain gradient lower bounds},
  author={Jordan T. Ash and Chicheng Zhang and Akshay Krishnamurthy and John Langford and Alekh Agarwal},
  journal={ICLR},
  year={2020}
}

@inproceedings{emoc2014frey,
  title={Selecting influential examples: Active learning with expected model output changes},
  author={Freytag, Alexander and Rodner, Erik and Denzler, Joachim},
  booktitle={European conference on computer vision},
  pages={562--577},
  year={2014},
  organization={Springer}
}

@inproceedings{emoc2016kading,
  title={Active and Continuous Exploration with Deep Neural Networks and Expected Model Output Changes},
  author={K{\"a}ding, Christoph and Rodner, Erik and Freytag, Alexander and Denzler, Joachim},
  booktitle={NIPSW},
  year={2016}
}

@inproceedings{emoc_reg2018kading,
  title={Active Learning for Regression Tasks with Expected Model Output Changes.},
  author={K{\"a}ding, Christoph and Rodner, Erik and Freytag, Alexander and Mothes, Oliver and Barz, Bj{\"o}rn and Denzler, Joachim and AG, Carl Zeiss},
  booktitle={BMVC},
  pages={103},
  year={2018}
}

@inproceedings{lookahead2022mohamadi,
  title={Making look-ahead active learning strategies feasible with neural tangent kernels},
  author={Mohamadi, Mohamad Amin and Bae, Wonho and Sutherland, Danica J},
  booktitle={NeurIPS},
  volume={35},
  pages={12542--12553},
  year={2022}
}

@inproceedings{sener2017active,
title={Active Learning for Convolutional Neural Networks: A Core-Set Approach},
author={Ozan Sener and Silvio Savarese},
booktitle={ICLR},
year={2018}
}

@inproceedings{typiclust2022hacohen,
  title={Active Learning on a Budget: Opposite Strategies Suit High and Low Budgets},
  author={Hacohen, Guy and Dekel, Avihu and Weinshall, Daphna},
  booktitle={Proceedings of the 39th International Conference on Machine Learning (ICML)},
  pages={8175--8195},
  year={2022},
  publisher={PMLR},
}

@inproceedings{probcover2022yehuda,
  title={Active learning through a covering lens},
  author={Ofer Yehuda and Avihu Dekel and Guy Hacohen and Daphna Weinshall},
  booktitle={NeurIPS},
  year={2022}
}

@inproceedings{bae2024maxherding,
  title={Generalized Coverage for More Robust Low-Budget Active Learning},
  author={Bae, Wonho and Noh, Junhyug and Sutherland, Danica J},
  booktitle={European Conference on Computer Vision},
  pages={318--334},
  year={2024},
  organization={Springer}
}

@inproceedings{uherding2024base,
  title={Uncertainty Herding: One Active Learning Method for All Label Budgets},
  author={Bae, Wonho and Oliveira, Gabriel L and Sutherland, Danica J},
  booktitle={ICLR},
  year={2025},
}

@article{houlsby2011bald,
  title={Bayesian active learning for classification and preference learning},
  author={Houlsby, Neil and Husz{\'a}r, Ferenc and Ghahramani, Zoubin and Lengyel, M{\'a}t{\'e}},
  journal={arXiv preprint arXiv:1112.5745},
  year={2011}
}

@article{kendall2017uncertainties,
  title={What uncertainties do we need in bayesian deep learning for computer vision?},
  author={Kendall, Alex and Gal, Yarin},
  journal={Advances in neural information processing systems},
  volume={30},
  year={2017}
}

@inproceedings{yoo2019learning,
  title={Learning loss for active learning},
  author={Yoo, Donggeun and Kweon, In So},
  booktitle={Proceedings of the IEEE/CVF conference on computer vision and pattern recognition},
  pages={93--102},
  year={2019}
}

@inproceedings{ma2024breaking,
  title={Breaking the barrier: Selective uncertainty-based active learning for medical image segmentation},
  author={Ma, Siteng and Wu, Haochang and Lawlor, Aonghus and Dong, Ruihai},
  booktitle={ICASSP 2024-2024 IEEE International Conference on Acoustics, Speech and Signal Processing (ICASSP)},
  pages={1531--1535},
  year={2024},
  organization={IEEE}
}

@inproceedings{mottaghi2014context,
  title={The Role of Context for Object Detection and Semantic Segmentation in the Wild},
  author={Mottaghi, Roozbeh and Chen, Xianjie and Liu, Xiaobai and Cho, Nam-Gyu and Lee, Seong-Whan and Fidler, Sanja and Urtasun, Raquel and Yuille, Alan},
  booktitle={Proceedings of the IEEE Conference on Computer Vision and Pattern Recognition},
  pages={891--898},
  year={2014}
}

@article{bansal2017pixelnet,
  title={Pixelnet: Towards a general pixel-level architecture},
  author={Bansal, Aayush and Chen, Xinlei and Russell, Bryan and Gupta, Abhinav and Ramanan, Deva},
  journal={arXiv preprint arXiv:1609.06694},
  year={2017}
}

@inproceedings{gu2020cagnet,
  title={CaGNet: Context-aware Feature Generation for Zero-shot Semantic Segmentation},
  author={Gu, Zhangxuan and Lin, Hanwang and Koniusz, Piotr and Li, Hong and Yang, Ming and Qiao, Yu and Fu, Yanwei},
  booktitle={Proceedings of the 28th ACM International Conference on Multimedia},
  pages={1921--1929},
  year={2020}
}

@inproceedings{kirillov2023segment,
  title={Segment anything},
  author={Kirillov, Alexander and Mintun, Eric and Ravi, Nikhila and Mao, Hanzi and Rolland, Chloe and Gustafson, Laura and Xiao, Tete and Whitehead, Spencer and Berg, Alexander C and Lo, Wan-Yen and others},
  booktitle={International Conference on Computer Vision},
  pages={4015--4026},
  year={2023}
}
